\documentclass[twoside,11pt]{article}

\usepackage{blindtext}
\usepackage{jmlr2e}
\usepackage{lastpage}

\usepackage{graphicx}
\usepackage{hyperref,amsfonts,latexsym,xspace, amsmath}
\usepackage{algorithm}
\usepackage{algpseudocode}
\usepackage[toc,page]{appendix}

\graphicspath{ {./plot/} }

\ShortHeadings{Predictability Analysis of Regression Problems via Conditional Entropy Estimations}{Fang and Lee}

\begin{document}
\title{Predictability Analysis of Regression Problems via Conditional Entropy Estimations}
\author{\name Yu-Hsueh Fang \email d11725001@ntu.edu.tw\\
       \addr Department of Information Management\\
       National Taiwan University\\
       \AND
       \name Chia-Yen Lee \email chiayenlee@ntu.edu.tw\\
       \addr Department of Information Management\\
       National Taiwan University}
\editor{}
\maketitle

\setlength{\baselineskip}{18pt}
\begin{abstract}
In the field of machine learning, regression problems are pivotal due to their ability to predict continuous outcomes.
Traditional error metrics like mean squared error, mean absolute error, and coefficient of determination measure model accuracy.
The model accuracy is the consequence of the selected model and the features, which blurs the analysis of contribution.
Predictability, in the other hand, focus on the predictable level of a target variable given a set of features.
This study introduces conditional entropy estimators to assess predictability in regression problems, bridging this gap.
We enhance and develop reliable conditional entropy estimators, particularly the KNIFE-P estimator and LMC-P estimator, which offer under- and over-estimation, providing a practical framework for predictability analysis.
Extensive experiments on synthesized and real-world datasets demonstrate the robustness and utility of these estimators.
Additionally, we extend the analysis to the coefficient of determination \(R^2 \), enhancing the interpretability of predictability.
The results highlight the effectiveness of KNIFE-P and LMC-P in capturing the achievable performance and limitations of feature sets, providing valuable tools in the development of regression models.
These indicators offer a robust framework for assessing the predictability for regression problems.
\end{abstract}
\begin{keywords}
  Predictability Analysis, Conditional Entropy Estimation, Information Theory, Regression Problem
\end{keywords}
\section{Introduction}
In the ever-evolving field of machine learning, regression problems hold a pivotal role due to their ability to model and predict continuous outcomes based on input data~\citep{FS15,MFD19}.
Traditionally, the objective of regression models is to minimize error metrics such as mean squared error (MSE) and mean absolute error (MAE).
The fitness of a model can also be assessed by metrics such as the coefficient of determination.
While these metrics provide insights into model accuracy, they fail to directly offer insights into how the features contribute to forecasting the target variable.
Predictability, in contrast, focuses on the degree to which a variable can be forecasted based on available features.

Predictability analysis has aided the study of complex systems such as transportation, physics, weather, human mobility, and human social behavior~\citep{KH05,CS10,TT11,VK19,GL22}.
However, there has been less study of predictability applied to machine learning tasks.
To bridge this gap, this study aims to provide an estimator and a systematic way to assess predictability with a specific focus on regression problems.

Information theory has gained significant traction in predictability analysis.
There are primarily two approaches to predictability analysis: information theory and chaos analysis. Chaos analysis has been successful in time series with few variables~\citep{AW85, AN01, PS05}.
However, it faces challenges when generalizing to multivariate analysis.
In contrast, the information theory-based approach benefits from advancements in machine learning and the study of underlying distributions, which has become the mainstream method for predictability analysis.
Similarly, information theory has been widely applied in machine learning.
Many studies suggest that metrics like mutual information are suitable for feature engineering, especially for capturing nonlinear and multivariate relationships~\citep{RB94,BF13,JRV14,JL17,XW20,DFV21}.

This study focuses on the estimation of conditional entropy and its use as an indicator of predictability.
The relationship between traditional error metrics and conditional entropy makes conditional entropy suitable for predictability analysis in supervised learning.
Conditional entropy, a measure derived from information theory, quantifies the uncertainty remaining about variables given other variables.
It also indicates the admissible region of prediction models, highlighting the achievable performance of regression models~\citep{BF13}.

In contrast to the developments of mutual information estimators, conditional entropy estimators are less frequently discussed.
The lack of a reliable estimator hinders the application of predictability analysis.
The core contribution of this study is the development of conditional entropy estimators for predictability analysis.
These estimators are designed to be reliable providing both over and lower estimations, which together give us a range for the true conditional entropy.
By establishing this range, the study offers a practical framework for analyzing predictability beyond conventional error metrics.

The reliability of our estimator is demonstrated through experiments on a synthesized dataset with known conditional entropy.
Moreover, we conduct experiments across various datasets with varying levels of information loss to illustrate the practical value of our estimator.
These experiments highlight the estimator's robustness and utility in real-world scenarios where data may be incomplete and the distribution of the noise is unknown.

By showcasing the performance of our conditional entropy estimator in both controlled and practical settings, this study underscores its relevance and applicability in assessing the predictability of regression models.
The insights gained from these experiments contribute to a deeper understanding of achievable model performance and the contribution of feature information in making predictions.

\section{Background}
The objective of a regression problem is to determine a prediction function \( f(x) \) given features \( X \in \mathbb{R}^d \) as input while minimizing the error between the predicted value and the label \( Y \in \mathbb{R} \)~\citep{FS15,MFD19}.
The dataset is assumed to have \( N \) instances.
We have the MSE $\epsilon_{MSE} = \sum^{N}_{i=1}(f(x_i)-y_i)^2/N$ and the mean absolute error (MAE) $\epsilon_{mae} = \sum^{N}_{i=1}|f(x_i)-y_i|/N$.
The training process of the prediction function \( f(X) \) often aims to minimize these two error criteria.
The search for \( f(X) \) includes time-consuming feature engineering, modeling, and parameter tuning. The existence of a measurable bound for MSE and MAE supports our ability to reject a feature set \( X \) that provides little information about \( Y \).
Hence, this study introduces predictability analysis for assessing the achievable MSE or MAE with the given features.

\subsection{Predictability and Conditional Entropy}
The connection between information theory measurements and predictability has driven numerous studies.
For instance, \cite{KH05} used a Gaussian ensemble prediction to assess relative entropy for predictability analysis.
\cite{CS10} studied the entropy of human mobility to demonstrate the limitations of predicting human actions.
\cite{TT11} evaluated random entropy, uncorrelated entropy, and conditional entropy to study the uncertainty of conversation. 
More recently, \cite{GL22} applied conditional entropy to traffic forecasting, highlighting the limitations of traffic predictability.
They also conclude that most prediction models fall into two categories: deterministic and probabilistic.
Deterministic models, such as those used in regression problems, provide a point-to-point mapping, giving the mean of the target probability density function (PDF).
On the other hand, probabilistic models capture the entire PDF, which can assist in the study or construction of deterministic models by providing a more comprehensive understanding of the uncertainty in the predictions.

Among all the information theory measurements, we focus on conditional entropy as our predictability indicator due to its intuitive relationship with MSE and MAE.
\cite{BF13} elaborates on the relationship between conditional entropy \( H(Y|X) \) and these error criteria.
Under the three commonly assumed noise distributions, Table~\ref{table:relationship} illustrates these relationships.
These relationships demonstrate the capability of conditional entropy as an indicator of the predictability of \( Y \) given \( X \).
Equality holds when the prediction function \( f(x) \) is perfect, in which case the error is merely the noise (\( H(Y|X) = H(f(X) + \epsilon | X) = H(\epsilon | X) = H(\epsilon) \)).
\begin{table}
  \caption{The relationship between error $\epsilon$ and conditional entropy $H(Y|X)$ \label{table:relationship}}
  \begin{center}
    \begin{tabular}{ c c c }
      \hline 
      Noise Distribution & Relationship with MSE & Relationship with MAE \\
      \hline
      \rule{0pt}{2.5ex}
      Uniform $\mathcal{U}(w,\Delta)$ & $\epsilon_{MSE}\geq\frac{1}{12}e^{2H(Y|X)}$ & $\epsilon_{mae}\geq\frac{1}{4}e^{2H(Y|X)}$ \\
      \rule{0pt}{2.5ex}
      Laplacian $\mathcal{L}(w,\lambda)$ & $\epsilon_{MSE}\geq\frac{1}{2e^2}e^{2H(Y|X)}$ & $\epsilon_{mae}\geq\frac{1}{2e}e^{2H(Y|X)}$ \\
      \rule{0pt}{2.5ex}
      Gaussian $\mathcal{N}(w,\sigma)$  & $\epsilon_{MSE}\geq\frac{1}{2\pi e}e^{2H(Y|X)}$ & $\epsilon_{mae}\geq\frac{1}{\pi \sqrt{e}}e^{2H(Y|X)}$ \\
      \hline 
     \end{tabular}
  \end{center}
\end{table}
If the assumption of the noise distribution is given, we can bound the best possible error using the estimated conditional entropy.
Otherwise, the lowest bounds among these three noise distributions are Gaussian for MSE and Laplacian for MAE.

\begin{figure}[h]
  \includegraphics[width=\textwidth]{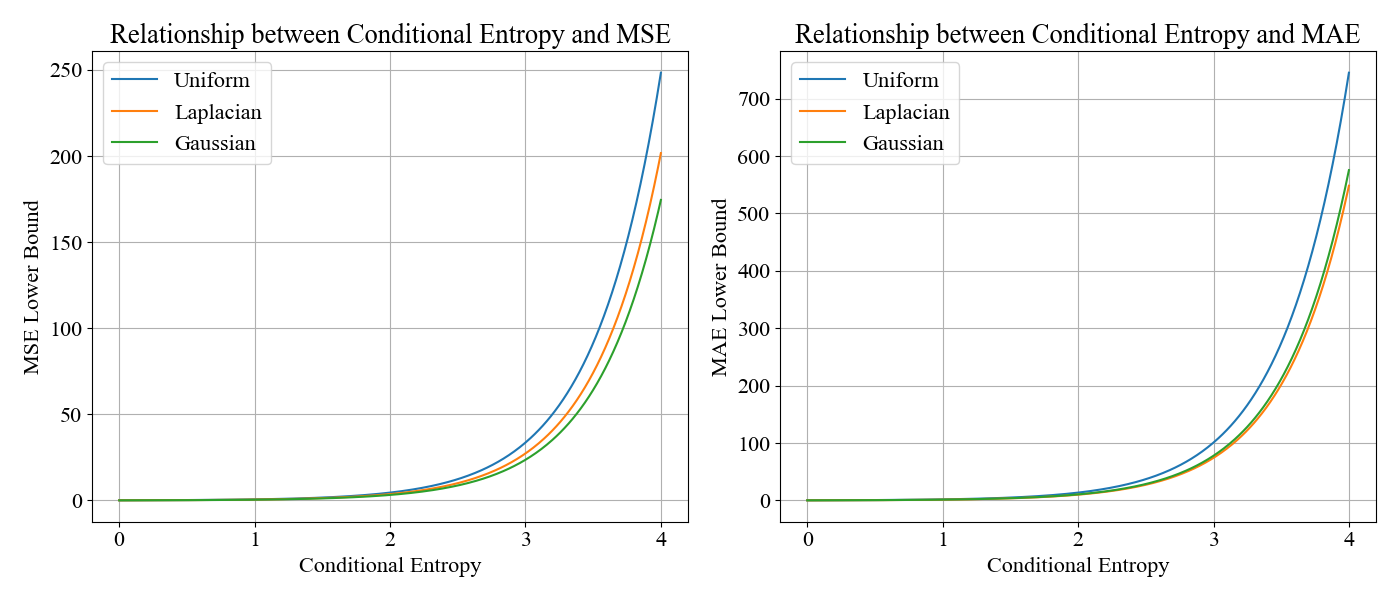}
  \caption{The admissible region shown by conditional entropy $H(Y|X)$ \citep{BF13}\label{fig-relationship}}
\end{figure}
Several challenges hinder the application of conditional entropy in predictability analysis. Firstly, the focus of estimator studies has traditionally been on mutual information \( I(X; Y) \) or differential entropy \( H(X) \), rather than conditional entropy \( H(Y|X) \).
Although the development of these estimators can benefit each other, conditional entropy estimators are less thoroughly investigated. 

Moreover, most existing conditional entropy estimators tend to provide over estimates.
As shown in Figure~\ref{fig-relationship}, conditional entropy serves as the lower bound of the error criteria.
Over estimates of conditional entropy do not necessarily bound the achievable error.
To address these issues, this study focuses on developing conditional entropy estimators that provide over and under estimations.
By doing so, we aim to offer a practical framework for evaluating predictability in regression problems, enabling better assessment of the achievable performance.

\subsection{Information Theory and Estimator}
Information theory, originally introduced by~\cite{CES48}, is a study of uncertainty in a system.
For two continuous random variables $X$ and $Y$ with joint probability density function $p(y,x)$, marginal probability density functions $p(x)$ and $p(y)$, and the conditional density function $p(y|x)$, we introduce the differential measurements.
The entropy $H(X)$ quantifies the amount of information of a random variable $X$.
\begin{equation}
H(X) = -\mathbb{E}_{p(x)}[ \log p(x) ]
\end{equation}
Mutual information $I(X;Y)$ measures the amount of information shared between two random variables. 
\begin{equation}
I(X; Y) = \mathbb{E}_{p(x,y)}\left[\log(\frac{p(x, y)}{p(x) p(y)})\right]
\end{equation}
It is symmetric, hence $I(X;Y)=I(Y;X)$.
Conditional entropy $H(Y|X)$ measures the uncertainty remaining in a target variable $Y$ given the value of another feature variable $X$. 
\begin{equation}
H(Y|X) = -\mathbb{E}_{p(x,y)}[\log p(y|x)]
\end{equation}
Mutual information can be expressed as the difference between the entropy of a variable and its conditional entropy given another variable, reflecting the amount of information shared between the variables.
\begin{equation}\label{eq:doe}
  I(X;Y)=H(Y)-H(Y|X)
\end{equation}
Equation~\ref{eq:doe} shows the compatibility of one estimators to other measurements.

The estimation of mutual information is an extensively studied topic.
One popular approach for estimating mutual information is through dual representations, which provide lower bounds for the estimation process
Based on \( f \)-divergence, several estimators have been proposed~\citep{XLN10,AO18}.
Specifically, \cite{AO18} proposes the estimator \( I_{InfoNCE} \), which utilizes the \( f \)-divergence representation and has become foundational in contrastive learning:
\begin{equation}
  I_{InfoNCE} := \mathbb{E}_{p(x,y)} \left[ \log \left( \frac{e^{f(x,y)}}{\frac{1}{N} \sum_{j=1}^{N} e^{f(x,y_j)}} \right) \right]
\end{equation}

The well-known Kullback-Leibler (KL) divergence is also a popular basis for estimators~\citep{SK97}.
\cite{MIB18} utilizes the dual representation of KL-divergence to provide the mutual information neural estimator \( I_{MINE} \):
\begin{equation}
  I_{MINE} := \mathbb{E}_{p(x,y)}[f(x,y)] - \log(\mathbb{E}_{p(x)p(y)}[e^{f(x,y)}])
\end{equation}

\cite{DM20} shows that $I_{MINE}$ is infeasible to give high-confidence lowerbound gaurantees for large mutual information.
To find an unbiased estimator,~\cite{DM20} proposed the difference-of-entropies (DoE) method.
The DoE method utilizes Equation~\ref{eq:doe} and the relationship between cross entropy and entropy.
\begin{equation}
  -\sum_{x\in X} p(x)\log q(x) \geq -\sum_{x\in X} p(x) \log p(x) = H(X)
\end{equation}
The equality holds when $p(x)=q(x)$.
By minimizing cross entropy, they estimate entropy with approximated probability density function $q(y)$ and $q(y|x)$:
\begin{align}
  &I_{DoE}=H_{q}(Y)-H_{q}(Y|X)\\
  &H_{q}(Y)=-\frac{1}{N}\sum_{i=1}^{N}\ln(q(y_i))\\
  &H_{q}(Y|X)=-\frac{1}{N}\sum_{i=1}^{N}\ln(q(y_i|x_i))
\end{align}
\cite{DM20} parameterizes $q(y_i|x_i)$ by Gaussian distribution or logistic distribution.
To make it more general,~\cite{GP22} proposed a kernelized-neural differential entropy estimation (KNIFE) with variational distribution $q_{\theta}(x)$ and $q_{\theta}(y|x)$.
To approximate marginal distribution $p(x)$, parameter $\theta$ includes kernel weights $w_k$, standard deviations $\sigma_k$, and the mean of the $k$ kernel $\mu_k$.
\begin{equation}
  q_{\theta}(x)=\sum_{k=1}^{K}w_k(\frac{1}{\sigma_k\sqrt{2\pi}}e^{-\frac{1}{2}(\frac{x-\mu_k}{\sigma_k})^2})
\end{equation}
Paramter $K$ denotes the number of kernels used.
If it is a conditional density estimation, then the neural network $\theta$ takes $x\in X$ as input and serves as parameter functions.
\begin{equation}
  q_\theta(y|x)=q(y;\theta(x))=\sum_{k=1}^{K}w_k(\theta)(\frac{1}{\sigma_k(\theta)\sqrt{2\pi}}e^{-\frac{1}{2}(\frac{y-\mu_k(\theta)}{\sigma_k(\theta)})^2})
\end{equation}
It does not require all parameters controlled by a neural network or trainable.
For instance, we can set the means of Gaussian distributions to instances of $y$ in training datasets, and then the Gaussian mixture model becomes a kernelized-neural density estimation.
Generally, the key of this approach is the density estimation, which can be enhanced by normalization and perturbation~\citep{JR19,JR20}.

To be a reliable indicator for predictability analysis, the characteristic of the estimator is crucial.
Unlike other approaches that focus on finding lower bounds \cite{PC20} proposes a contrastive log-ratio upper bound (CLUB) to approximate the upper bound of mutual information:
\begin{equation}
  I_{CLUB} := \mathbb{E}_{p(x,y)}[\log q(y|x)] - \mathbb{E}_{p(x)p(y)}[\log q(y|x)]
\end{equation}

To estimate conditional entropy, we can use any \( H(Y) \) estimator with mutual information estimators.
More intuitively, we can estimate the conditional density function directly.
However, these conditional entropy estimators are theoretically upperbounds (minimizing cross entropy) or non-bias estimators, which can limit their insights for predictability analysis.
This study aims to address these issues by developing more reliable conditional entropy estimators, offering both over and under estimations, to enhance the practical analysis of predictability in regression problems.

\section{Conditional Entropy Under Estimator}
The absence of a reliable conditional entropy under estimator prevents us from a comprehensive predictability analysis.
For composing a conditional entropy under estimator, we use the state-of-art mutual information upperbound as the foundation.
Assuming $p(y)$ is given, here is a conditional entropy lowerbound using CLUB.
\begin{align*}
  H(Y)-I_{CLUB}(X;Y) &=-\mathbb{E}_{p(y)}[\log p(y)]-\mathbb{E}_{p(x,y)}[\log q_\theta(y|x)]+\mathbb{E}_{p(x)p(y)}[\log q_\theta(y|x)].
\end{align*}
Since $I_{CLUB}(X;Y)\geq I(X;Y)$, $H(Y|X)\geq H(Y)-I_{CLUB}(X;Y)$.
Unfortunatly, $p(y)$ is unknown in most applications.
Then, one may approximate $p(y)$ by sampling and $q_\theta(y|x)$.
The conditional entropy lowerbound based on CLUB is defined as
\begin{align*}
  H_{CLUB}(Y|X) &=-\mathbb{E}_{p(x)}[\log \mathbb{E}_{p(y)}[q_\theta(y|x)]]-\mathbb{E}_{p(x,y)}[\log q_\theta(y|x)]+\mathbb{E}_{p(x)}\mathbb{E}_{p(y)}[\log q_\theta(y|x)].
\end{align*}
Having a set of data with label $Y=\{y_1,\dots,y_N\}$ and features $X=\{x_1,\dots,x_N\}$, $H_{CLUB}$ has an estimator
\[
  \hat{H}_{CLUB}(Y|X;\theta) =-\sum_{i=1}^{N}[\log(\frac{1}{N}\sum_{j=1}^{N}[q_\theta(y_i|x_j)])+\log q_\theta(y_i|x_i)-\log {q_\theta(y_i|\tilde{x}_i)}].
\]
Variables $\tilde{y}_i$ are randomly drawn from $y_i$.
The time complexity of using $\hat{H}_{CLUB}$ is $\mathcal{O}(N^2)$ which lowers its pratical utility.
Also, it is difficult to balance the fit to the conditional density function between its approximation and the marginal distribution.

\subsection{Conditional Entropy Lowerbound of Marginal Density Function and Conditional Density Function}
As a complement, we derive a new conditional entropy lowerbound based on a pair of estimated density functions.
The proposed conditional entropy lowerbound of marginal density function and conditional density function (LMC) is written as
\[
  H_{LMC}(Y|X;\theta) =-\mathbb{E}_{p(y)}[\log {q_\theta(y)}]-\mathbb{E}_{p(x,y)}[\log q_{\theta}(y|x)]+\mathbb{E}_{p(x)p(y)}[\log q_\theta(y|x)].
\]
The variational density functions $q_\theta(y|x)$ and $q_\theta(y)$ are estimated density functions of conditional density function $p(y|x)$ and marginal density function $p(y)$.
Having a set of data with label $Y$ and features $X$, $H_{LMC}$ has an estimator
\[
  \hat{H}_{LMC}(Y|X;\theta) =\frac{1}{N}\sum_{i=1}^{N}[-\log {q_\theta(y_i)}-\log q_{\theta}(y_i|x_i)+\log q_\theta(y_i|\tilde{x}_i)].
\]
LMC takes advantage of a contrastive form and Jensen's inequality as CLUB, but complements the deficiencies estimating conditional entropy.

We can replace $q_\theta(y|x)$ in LMC by an estimated joint density function $q_\theta(x,y)$.
Then, we have a conditional entropy lowerbound of marginal density function and joint density function (LMJ)
\[
  H_{LMJ}(Y|X;\theta) =-\mathbb{E}_{p(y)}[\log {q_\theta(y)}]-\mathbb{E}_{p(x,y)}[\log q_{\theta}(x,y)]+\mathbb{E}_{p(x)p(y)}[\log q_\theta(x,y)].
\]
Assuming $q_{\theta}(x,y)/p(x)=q_\theta(y|x)$, one can found that LMJ is identical to LMC.
But, LMC has the advantage of not only estimating lowerbound of conditional entropy, but also upperbound.
As a consequence, we focus our discussion on LMC.

\subsection{Gaps between LMC and Conditional Entropy}
We examine how the additional marginal density estimation \( q_\theta(y) \) in LMC impacts gaps.
The following derivations generally follow the CLUB estimator~\citep{PC20}, but incorporate two estimations.
To ensure the self-consistency, we derive the following theory independently. First, we consider the general gap between LMC and conditional entropy.
\begin{align*}
  \Delta_{LMC} =&H(Y|X)-H_{LMC}(Y|X;\theta)\\
           =&-\mathbb{E}_{p(x,y)}[\log p(y|x)]+\mathbb{E}_{p(y)}[\log {q_\theta(y)}]+\mathbb{E}_{p(x,y)}[\log q_{\theta}(y|x)]-\mathbb{E}_{p(x)p(y)}[\log q_\theta(y|x)]\\
           =&\mathbb{E}_{p(x)p(y)}[\log \dfrac{p (y)}{q_\theta (y|x)}]-\mathbb{E}_{p(x,y)}[\log \dfrac{p(y|x)}{q_\theta (y|x)}]-\mathbb{E}_{p(y)}[\log \dfrac{p (y)}{ q_\theta(y)}]\\
           =&\mathbb{E}_{p(x)p(y)}[\log \dfrac{p(x)p(y)}{q_\theta (y|x)p(x)}]-\mathbb{E}_{p(x,y)}[\log \dfrac{p(y|x)}{q_\theta (y|x)}]-\mathbb{E}_{p(y)}[\log \dfrac{p (y)}{ q_\theta(y)}]\\
           =&KL(p(x)p(y)||q_\theta (y|x)p(x))-(KL(p(y|x)||q_\theta(y|x))+KL(p(y)||q_\theta(y)))
\end{align*}
Three Kullback-Liebler (KL) divergence compose $\Delta_{LMC}$.
The first two KL divergence are the general gap of $I_{CLUB}$ which has been shown to be non-negative with proper optimization.
The last KL divergence comes from the extra marginal density estimation.
LMC is a conditional entropy lowerbound with non-negative $\Delta_{LMC}$
\begin{theorem}\label{Theorem:GeneralGap}
 $H_{LMC}(Y|X,\theta)$ is a conditional entropy lowerbound under
  \[KL(p(x)p(y)||q_\theta (y|x)p(x))>KL(p(y|x)||q_\theta(y|x))+KL(p(y)||q_\theta(y)).\]
\end{theorem}
Theorem~\ref{Theorem:GeneralGap} implies how to optimize LMC.
\cite{PC20} has shown that minimizing cross entropy of two density function also minimizes their KL divergence.
Hence, we use two cross entropy loss to train the variational density functions.
\begin{align}
  &\min_{\theta} -\frac{1}{N}\sum_{i=1}^{N}(\log q_\theta(y_i|x_i)) \label{eq:cond_loss}\\
  &\min_{\theta} -\frac{1}{N}\sum_{i=1}^{N}(\log q_\theta(y_i))\label{eq:marg_loss}
\end{align}
Equation~\ref{eq:cond_loss} is the same objective function for train Gaussian mixture in \cite{GP22} which is an over estimator of $H(Y|X)$.
Similarly, we train $q_\theta(y)$ with Equation~\ref{eq:marg_loss} which approximates marignal entropy $H(Y)$.
As an application for supervised learning, the dimension of $y$ is rather small, which alleviates the burden of optimization.
Though fulfilling the lowerbound condition is simple, we still refer to the proposed method as an under estimator.

When $q_\theta(x|y)=p(x|y)$ and $q_\theta(y)=p(y)$, LMC has underlining density functions.
LMC with underlining density functions is denoted as $H^*_{LMC}$.
The gap between $H^*_{LMC}$ and the true conditional entropy is the optimal gap $\Delta^*_{LMC}$, which shows that $H^*_{LMC}$ is a strict lowerbound.
\begin{align*}
  \Delta^*_{LMC} =&KL(p(x)p(y)||p(y|x)p(x))-KL(p(y|x)||p(y|x))-KL(p(y)||p(y))\\
          =&KL(p(x)p(y)||p(x,y))\geq 0
\end{align*}
Since KL divergence is non-negative, $\Delta^*_{LMC}$ is non-negative.
The equality holds when $x$ is independent to $y$.
\begin{theorem}\label{Theorem:OptGap}
  For two random variables X and Y, $H^*_{LMC}(Y|X)$ is a lowerbound of conditional entropy $H(Y|X)$ with gap
  \[
     \Delta^*_{LMC}=KL(p(x)p(y)||p(x,y)).
  \]
\end{theorem}
Theorem~\ref{Theorem:OptGap} shows LMC is a conditional entropy lowerbound when the optimized.
Also, it supports applications which the density functions have been known.
In practice, we implement the variational distribution \( q_\theta(y) \) and \( q_\theta(y|x) \) as KNIFE estimator~\citep{GP22}.

Thanks to their universal approximation property, trainable Gaussian mixtures have the capacity to approximate any distribution~\citep{VM96, MFH08}.
Similarly, the high expressiveness of neural networks enables them to provide accurate approximations of \( q_\theta(y|x) \) to \( p(y|x) \) with appropriate network settings~\cite{ZL17,MR17,SO19}.
As a result, the divergences \( KL(p(y)||q_\theta(y)) \) and \( KL(p(y|x)||q_\theta(y|x)) \) can be arbitrarily small.
Theoretically, the LMC estimator is an underestimator of conditional entropy, leveraging the powerful approximation capabilities of Gaussian mixtures and neural networks.

\section{Estimating Conditional Entropy}
Before applying LMC as an indicator of predictability, we first validate its performance and address a few deficiencies.
An ideal conditional entropy estimator should capture nonlinearity, multivariate interactions, and variable dependencies. 
However, the high expressiveness of the estimator can easily lead to overfitting.
To mitigate overfitting, we augment the optimization process with perturbation techniques.
These techniques help to ensure that the estimator generalizes well to unseen data, maintaining its reliability as a conditional entropy estimator and, consequently, as an indicator of predictability.

\subsection{Normalization and Perturbation}
We apply two enhancements to stabilize the optimization process.
The first one is the z-score normalization of features.
Z-score normalization ensures each feature has a mean of zero and a standard deviation of one.
The formula for z-score normalization of a feature is given by:
\[
  z^x = \frac{x-\mu_x}{\sigma_x}
\]
where $\mu_x$ and $\sigma_x$ is the mean and the standard deviation of the feature values.
This process is crucial as it mitigates the problem of different scales across input features, which can lead to biased or inefficient learning processes.
Standardization makes gradient-based optimization more stable by preventing any single feature from dominating the update steps due to its scale \citep{JR19}.

Perturbation is another enhancement we applied to our estimator.
Different from the collected noise $\epsilon$, perturbation is a small random vector $\xi \sim p_\xi$ with standard deviation $\eta$ fulfilling
\[
    \mathbb{E}_{p_\xi}[\xi]=0\ \text{and}\ \mathbb{E}_{p_\xi}[\xi\xi^T]=\eta^2I.
\]
The perturbation augments features and target.
One popular choice of $p_\xi$ is the standard normal distribution
\[
    \phi(z)=\frac{e^{-\frac{z^2}{2}}}{\sqrt{2\pi}}.
\]
The standard deviation is controlled by a bandwidth $h$.
Generally, $h\rightarrow 0$ as the number of instance $N\rightarrow\infty$.
Hence, $h= \sigma N^{-1/5}$ is a popular choice.
Then, with $\sigma_z = 1$, we augment the normalized features $z^x$ and the target $y$ by
\[
  \tilde{y}_i = y_i+\sigma_yN^{-\frac{1}{5}}\xi^y\ \text{and}\ \tilde{z}^x_i = z^x+N^{-\frac{1}{5}}\xi^x.
\]
Perturbations $\xi^y$ and $\xi^x$ are drawn from $\phi(z)$.
\cite{JR20} have analyzed how perturbation affects optimization.
In short, adding perturbation is a smoothness regularization that penalizes large negative second derivatives w.r.t. features $x$.
\cite{JR20} show the impact of the perturbation with a real-world dataset.
We contribute a simulated dataset for validation.
The simulated dataset supports the true conditional entropy and differential entropy for better comprehension of optimization.

We summarize the proposed procedure in Algorithm~\ref{alg:lmc}.
The returned value \(\hat{H}^{lmcp}\) is the estimated LMC, while \(\hat{H}^{knifecp}\) is the overestimation of conditional entropy using the perturbed KNIFE estimator \( q_\theta(y|x) \).
Additionally, \(\hat{H}^{knifedp}\) is the overestimation of differential entropy using \( q_\theta(y) \) with perturbed training instances.
The overestimation of conditional entropy \(\hat{H}^{knifecp}\) is similar to the KNIFE estimator but complemented with normalization and perturbation.
Moreover, the overestimation of differential entropy \(\hat{H}^{knifedp}\) provides a baseline measurement of the amount of information needed to predict the target.
\begin{algorithm}
  \caption{LMC with perturbation and normalization}\label{alg:lmc}
  \begin{algorithmic}
  \Require $x,y,b,N$
  \State $\theta^c,\theta^m \gets$Initial neural network parameter
  \State $z^x \gets (x-\bar{x})/\sigma_x$
  \State $y_{train},y_{val},z^x_{train},z^x_{val}\gets$Seperate validation data and training data
  \While{Not converged}
  \State Draw a batch $y_i,z^x_i$ with size $b$ from $y_{train},z^x_{train}$
  \State Draw perturbation $\xi^y,\xi^x$ from $p_\xi$
  \State $\tilde{y}_i \gets y_i+\sigma_y b^{-1/5}\xi^y$
  \State $\tilde{z}^x_i \gets z^x_i+b^{-1/5}\xi^x$
  \State Evaluate loss functions:
  \State $H^c \gets -\frac{1}{b}\sum_{i=1}^{b}\log q_{\theta^c}(\tilde{y}_i|\tilde{z}^x_i)$
  \State $H^m \gets -\frac{1}{b}\sum_{i=1}^{b}\log q_{\theta^m}(\tilde{y}_i)$
  \State Update neural networks:
  \State $\theta^c \gets \theta^c+\nabla H^c$
  \State $\theta^m \gets \theta^m+\nabla H^m$
  \EndWhile
  \State Calculate entropies:
  \State $\hat{H}^{knifecp} \gets -\frac{1}{N}\sum\log q_{\theta^c}(y_{val}|z^x_{val})$
  \State $\hat{H}^{knifedp} \gets -\frac{1}{N}\sum\log q_{\theta^m}(y_{val})$
  \State $z^x_{shuffle} \gets$ Shuffle $z^x_{val}$
  \State $\hat{H}^{lmcp} \gets \hat{H}^{knifedp}+\hat{H}^{knifecp}+\frac{1}{N}\sum_{i=1}^{N}\log q_{\theta^c}(y_{val}|z^x_{shuffle})$\\
  \Return $\hat{H}^{knifecp},\hat{H}^{knifedp},\hat{H}^{lmcp}$
  \end{algorithmic}
\end{algorithm}
This algorithm ensures that the proposed LMC estimator is a reliable underestimator, mitigating the risk of overfitting through perturbation and proper optimization techniques.
Additionally, byproducts of this algorithm are the overestimations of conditional entropy and differential entropy, which further aid in the predictability analysis.

\subsection{Simulated Dataset}
In the pursuit of evaluating our conditional entropy estimator, it is essential to use datasets that reflect types of relationships among variables.
Table~\ref*{table:simfunc} illustrates four synthesized datasets, each embodying distinct functional characteristic.
\begin{table}
  \caption{Properties and task functions \label{table:simfunc}}
  \begin{center}
    \begin{tabular}{ l l l }
      \hline 
      Property & Function & Feature Set\\
      \hline
      Linearity & $y=x_1+\epsilon$ & $\{x_1\}$\\
      Nonlinearity & $y=50\sin(\pi x_1/50)+\epsilon$ & $\{x_1\}$ \\
      Interaction  &$y=x_1\sin(\pi x_2/50)+\epsilon$ & $\{x_1,x_2\}$\\ 
      Multivariate  &$y=x_1\sin(\pi x_2/50)+x_3+\epsilon$ & $\{x_1,x_2,x_3,x_4\}$\\ 
      \hline 
     \end{tabular}
  \end{center}

\end{table}
These datasets are designed to simulate different scenarios that might be encountered in real-world data analysis, thereby providing a comprehensive test suite for entropy estimation.

The first dataset is generated based on a simple linear relationship which is crucial for establishing the baseline.
Nonlinear dynamics are prevalent in many fields such as economics, biology, and physics, and they pose another challenge in entropy estimation.
The third dataset incorporates an interaction between two variables.
Interaction effects are critical in scenarios where the effects from variables are not independent of other variables.
This dataset helps in testing the estimator's ability to discern interactions between variables.
The final dataset extends the complexity by including multiple independent variables, which is more reflective of real-world situations where multiple factors simultaneously influence the outcome.
Each feature $x_i$ are drawn from the Uniform distribution with $\Delta=100$ and zero mean.

\begin{table}
  \caption{The relationship between the parameters and conditional entropy \label{table:relationship_parameters}}
  \begin{center}
    \begin{tabular}{ c c c }
      \hline 
      Noise Distribution & Relationship with Conditional Entropy \\
      \hline
      \rule{0pt}{2.5ex}
      Uniform $\mathcal{U}(w,\Delta)$ & $\ln(12\Delta)=H(Y|X)$ \\
      \rule{0pt}{2.5ex} 
      Laplacian $\mathcal{L}(w,\lambda)$ & $\ln(2e\lambda)=H(Y|X)$  \\
      \rule{0pt}{2.5ex}
      Gaussian $\mathcal{N}(w,\sigma)$ & $\ln(2\pi e\sigma^2)=H(Y|X)$ \\
      \hline 
     \end{tabular}
  \end{center}
\end{table}

We draw noise from three mentioned distribution.
Table~\ref*{table:relationship_parameters} summarizes the relationship between parameters and the conditional entropy.
Each type of noise distribution offers unique properties and challenges, which are crucial for testing the adaptability of entropy estimation.
Gaussian or normal noise is the most common type of noise encountered in real-world data.
With a probability density function that is sharper at the peak and decays exponentially, Laplacian noise is useful for modeling data with occasional large deviations.
This type of noise tests an estimator's ability to handle outliers and tailed distributions.
Uniform noise distributes all values equally within a specific range, introducing a scenario where the variation is constant.
This type of noise allows for the examination of an entropy estimator's performance in uniformly random environments.

\subsection{Mitigation of Overfitting}
The first experiment is designed to demonstrate the mitigation of overfitting through perturbation.
Overfitting typically arises when there is insufficient training data or when a model has excessively many parameters relative to the amount of input data.
To address these issues, we generate task instances with Gaussian error distribution with $\epsilon$ within the range $[250, 500, 1000]$.
The size of dataset $N$ is $3000$.
This variety allows us to evaluate the model's performance across different noise level.
The number of kernel $K$ is $600$.

We use two types of tasks to illustrate the overfitting: the linear task ($y=x_1+\epsilon$) and the interaction task ($y=x_1\sin(\pi x_2/50)+\epsilon$).
The linear task, being the simplest, is more prone to overfitting as it may not capture sufficient complexity from the data, leading to poor generalization on unseen data.
On the other hand, the nonlinear interaction task represents a more typical scenario with added complexity, which helps in understanding how well our perturbation methods can handle more realistic conditions.
These contrasting setups provide a comprehensive view of the effectiveness of perturbation in mitigating overfitting.

\begin{figure}[h]
  \includegraphics[width=\textwidth]{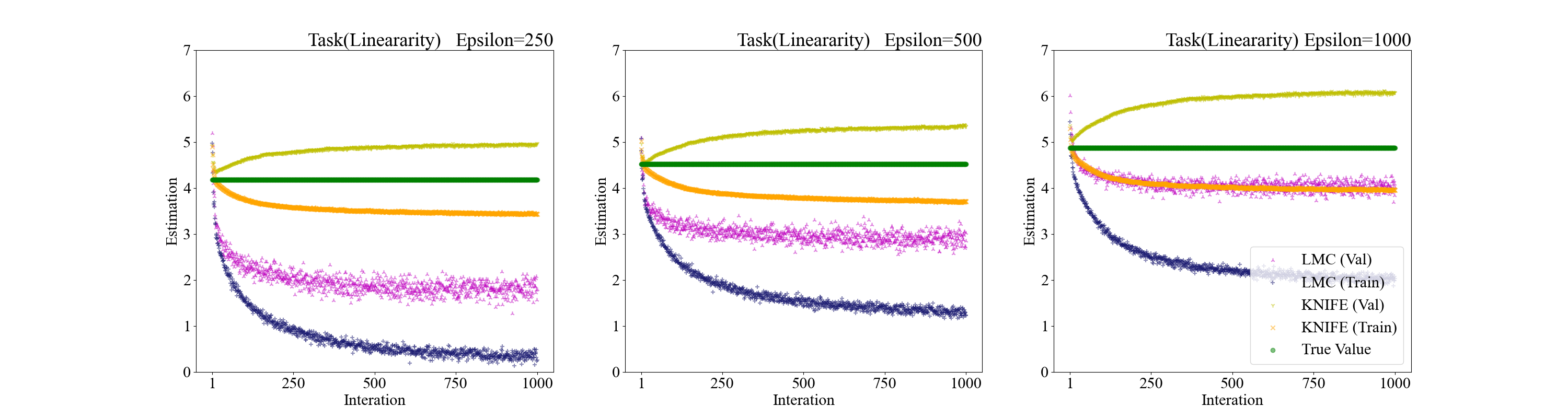}
  \caption{Fitting linear relationship without perturbation\label{fig-p0t0}}
\end{figure}

\begin{figure}[h]
  \includegraphics[width=\textwidth]{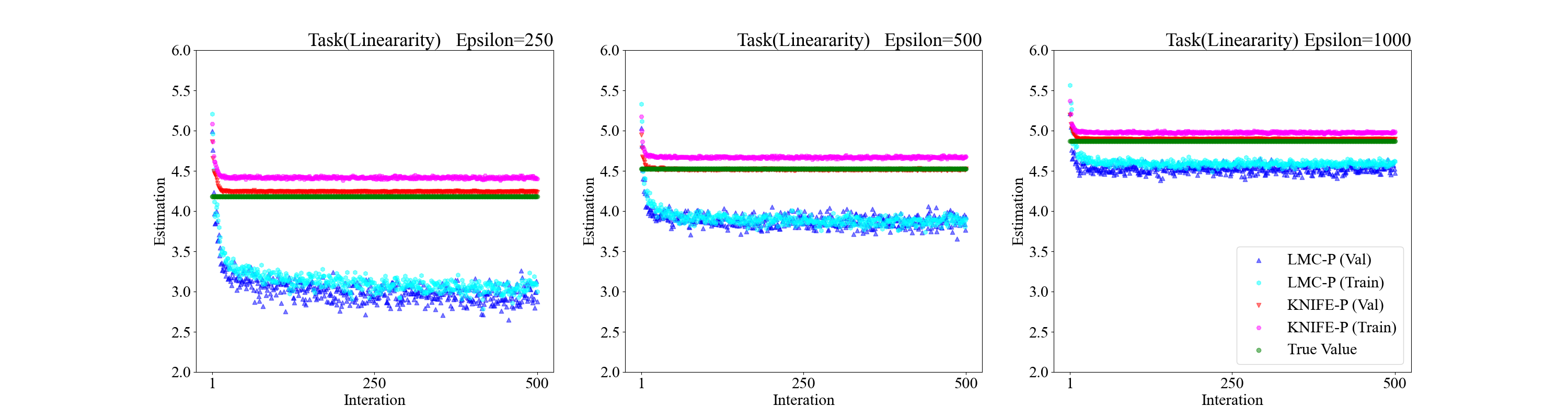}
  \caption{Fitting linear relationship with perturbation\label{fig-p1t0}}
\end{figure}

In Figure \ref{fig-p0t0} and Figure \ref{fig-p1t0}, we observe the fitting processes of the KNIFE and LMC models applied to a linear task.
The suffix "-P" indicates that the training process has been perturbed.
We present the convergence of conditional entropy estimation; hence, KNIFE-P represents the value of $\hat{H}^{knifecp}$ while LMC-P is $\hat{H}^{lmcp}$.
The occurrence of lower validation values compared to training values is attributed to the dropout in each layer.
The green line represents the true conditional entropy.


The results clearly indicate that without perturbation, KNIFE estimates significantly different values on the training set compared to the validation set.
In contrast, when perturbation is applied, KNIFE-P consistently provides similar values for both the training and validation sets, acting as an over estimator.
This suggests that perturbation enhances the robustness of the entropy estimate against underfitting.
Despite the simplicity of the task, KNIFE reliably maintains its over estimator characteristic when perturbation is applied during the training process.
This demonstrates that the perturbation technique effectively preserves the precision and reliability of the KNIFE estimator.

\begin{figure}[h]
  \includegraphics[width=\textwidth]{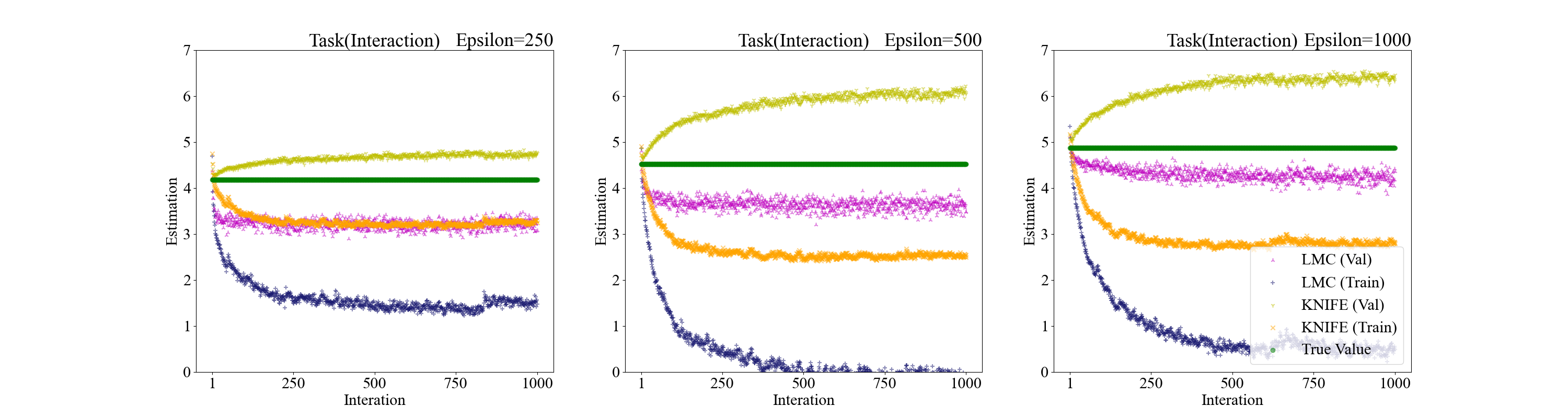}
  \caption{Fitting nonlinear interaction without perturbation\label{fig-p0t2}}
\end{figure}

\begin{figure}[h]
  \includegraphics[width=\textwidth]{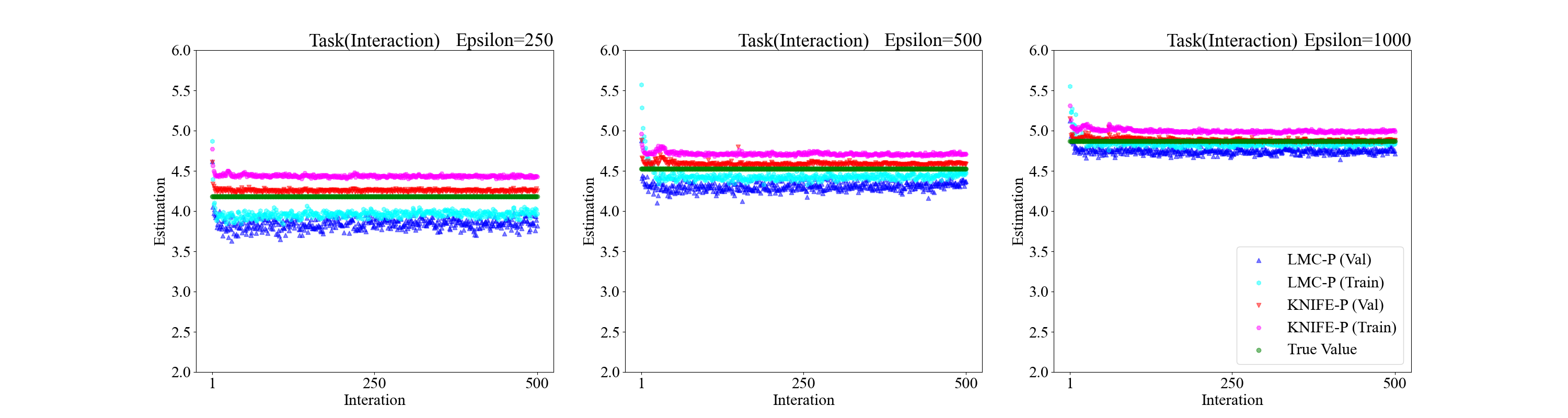}
  \caption{Fitting nonlinear interaction with perturbation\label{fig-p1t2}}
\end{figure}

In Figure \ref{fig-p0t2} and Figure \ref{fig-p1t2}, we observe the fitting processes of KNIFE applied to the nonlinear interaction task.
Both estimators successfully respond to the increasing true conditional entropy.
However, KNIFE fails to uphold its characteristic as an overestimator without perturbation.
Again, the estimated conditional entropy values differ significantly between the training set and the validation set.
However, when perturbation techniques are employed, the robustness of KNIFE significantly improves.
Perturbation not only mitigates overfitting, but also protects KNIFE's ability as an over estimator, which is vital for our application.
The figures clearly illustrate that with perturbation, both KNIFE and LMC are able to encompass the true conditional entropy within their estimated values, effectively capturing the boundaries.
While KNIFE maintains its feature as an over estimator with perturbation, LMC exhibit to serve reliably as an under estimator.
Capturing the boundaries of conditional entropy drives our application.

\subsection{Estimating MSE bounds}
\begin{figure}[p]
  \includegraphics[width=\textwidth]{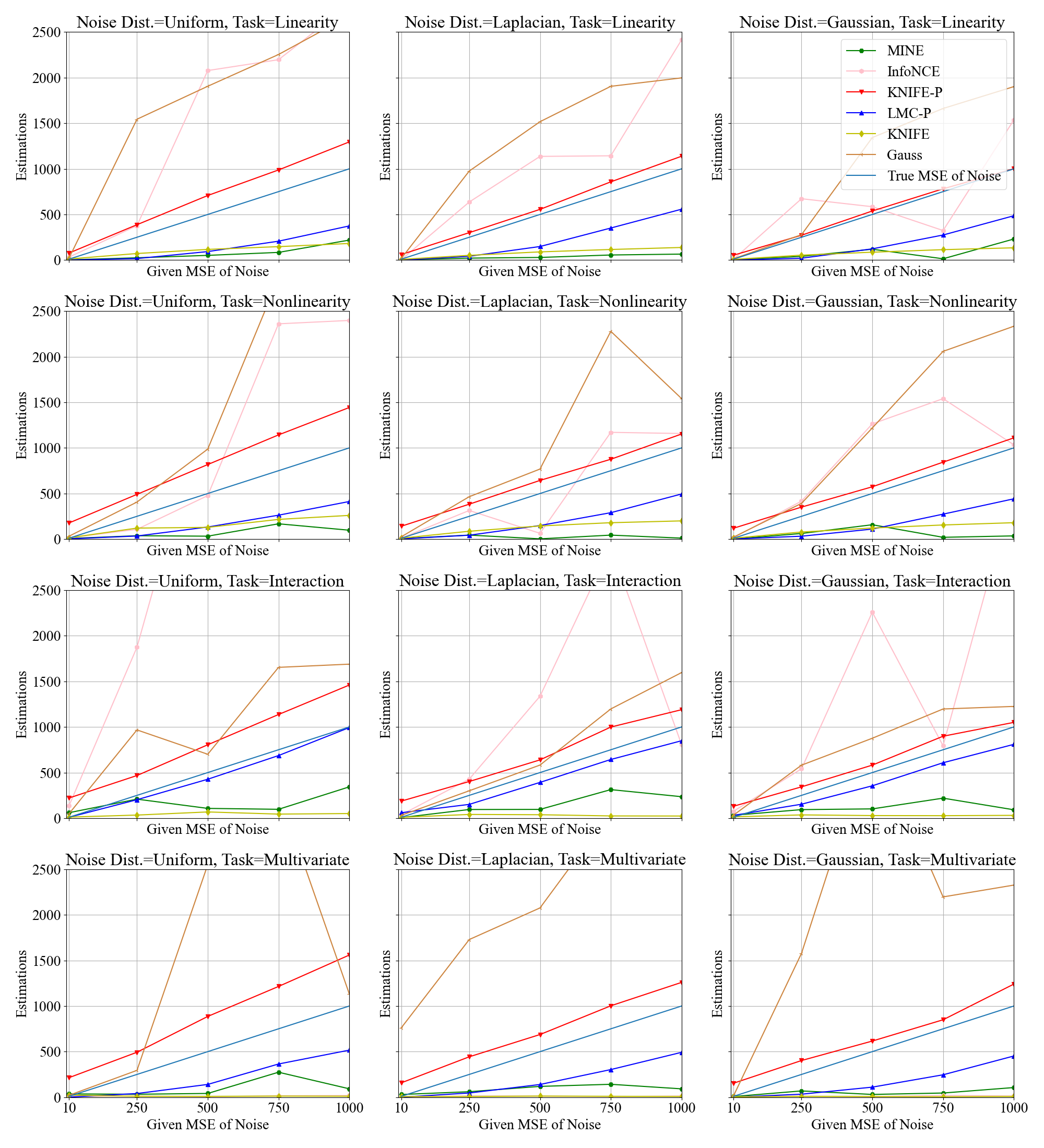}
  \caption{The comparison of different estimators\label{fig-syn}}
\end{figure}
The second experiment is designed to facilitate a comparative analysis of various estimators.
Based on DoE approach, we include an estimator Gauss using a Gaussian distribution as a benchmark.
We include mutual information estimators MINE and InfoNCE in this experiment, while CLUB is excluded due to its characteristic as a special biased estimator.

Mutual information possesses the intrinsic property that \( I(X; X) = H(X) \).
Therefore, we utilize the expression \( I(Y; Y) - I(X; Y) \) to estimate the conditional entropy \( H(Y | X) \) for MINE and InfoNCE.

For this experiment, we generate task instances with \( N = 3000 \) and incorporate three different types of noise, with MSE values spanning \( [10, 250, 500, 750, 1000] \).
This diversity in tasks is intended to assess the consistency and robustness of each estimator across varying levels of noise and complexity.

For KNIFE, KNIFE-P, and LMC-P, the number of kernels used is set to 600.
The dimensions of the hidden layer used, in order, are 16, 16, 256, and 256 for 4 tasks.
The number of layers are 4.
We undertake extensive training, conducting $1500$ epochs for each estimator to ensure adequate learning.
An early stopping mechanism is employed, which terminates training if the best estimation does not improve over $300$ epochs.
This approach prevents overfitting and optimizes computational efficiency.
To ensure reliability, each estimation represents the median outcome of $10$ repetitions.
The estimation variances are presented in the Appendix~\ref{append-syn-var}.

For a more intuitive understanding and practical relevance, the estimated medians of conditional entropy are converted to estimated MSE bounds using the equations provided in Table~\ref{table:relationship}.
This transformation aids in contextualizing the estimations within a more commonly understood metric, facilitating clearer comparisons and interpretations of the estimator performances.
However, this conversion also enlarges the error of the estimators, which should be taken into account when interpreting the results.

Figure~\ref{fig-syn} presents the summarized results of the comparative analysis between the different estimators.
The true MSE of the given noise is depicted by a plain line, serving as a benchmark against which the performance of the estimators can be evaluated.

The figure first includes results for two different types of tasks: Linearity tasks and Nonlinearity task from Table~\ref{table:simfunc}.
These categories are represented in the first two rows of the figure.
The results demonstrate that, under these straightforward criteria, most of the estimators are capable of providing estimated bounds that correlate appropriately with the increasing levels of noise.
This indicates that the estimators are generally effective in responding to changes in noise intensity.

In the analysis of the first two tasks, KNIFE-P demonstrates an average absolute error of $140.55$, effectively illustrating its capability to serve as a reference value for the MSE bound.
The tasks with Uniform noise are having the largest estimation error.
On the other hand, LMC-P, with an average absolute error of $322.80$, does not match the precision of KNIFE-P but retains its characteristic as an underestimator.
This attribute of LMC-P is significant as it establishes a strong lower boundary, indicating the admissible region for the prediction model.
MINE and KNIFE tend to provide lower estimations than the true value, while InfoNCE and Gauss yield higher estimations.

The third and fourth tasks, depicted in the last two rows of Figure~\ref{fig-syn}, are more complex and challenge the estimators further due to their increased complexity. 
For KNIFE-P, a larger error is inevitable in these tasks, with an average error of $220.96$.
This error is notably higher than that observed in the first two tasks, reflecting the additional challenges posed by the complexity of these tasks.
However, despite this increase in error, KNIFE-P still maintains its correlation with the increasing noise levels, affirming its robustness and utility even under more demanding conditions.
Notably, the Gauss estimator provides accurate estimations in the third task, suggesting possible developments based on such a simple approach.

LMC-P demonstrates notable consistency across different scenarios.
It achieves an average absolute error of $197.28$ in these last two tasks, indicating its reliable performance even as task complexity escalates.
This consistency is crucial for applications requiring dependable lower bounds across a range of conditions, highlighting LMC-P's strength in providing stable and consistent reference.
This performance further reinforces the utility of LMC-P as a reliable under estimator that can adapt to various scenarios.

In the majority of tasks, KNIFE-P and LMC-P successfully encompass the MSE line of noise, as illustrated in Figure~\ref{fig-syn}.
This observation underscores the capability of these two estimators to delineate the admissible region for a given dataset, effectively showcasing their practical utility.

KNIFE-P, in particular, serves as a reference bound for MSE, indicating a practical value that prediction models can realistically achieve.
This characteristic makes KNIFE-P especially valuable for setting realistic expectations and benchmarks in model performance, providing a gauge against which the efficiency of prediction models can be measured.

LMC-P acts as a strict bound, illustrating the limitations inherent in the set of features $X$ used for predicting the target $Y$.
This function of LMC-P is crucial for understanding the outer limits of what can be expected from a given feature set, thus helping in assessing the potential effectiveness and constraints of models before training prediction models.
Together, KNIFE-P and LMC-P offer complementary insights into the bounds of prediction accuracy, defining both the achievable potential and the fundamental limits of modeling.

We refrain from emphasizing the numerical error across all estimators due to their distinct purposes. InfoNCE is primarily designed to serve as an error criterion for optimization, and its main characteristic is its response to the level of noise.
MINE, on the other hand, is designed to estimate mutual information and also serve as an error criterion.
Their differing intents often lead to inevitable instability and inaccuracies in the estimation of conditional entropy.
Additionally, the exponential transformations detailed in Table~\ref{table:relationship} tend to amplify the errors of these estimations, further complicating direct comparisons.

This experiment aimed to evaluate the performance of several entropy estimators, including KNIFE-P, LMC-P, KNIFE, Gauss (DoE), MINE, and InfoNCE, across tasks of varying complexity and noise levels.
The tasks spanned simpler linear and nonlinear scenarios as well as more complex settings.
Most estimators effectively responded to increasing noise levels, with KNIFE-P and LMC-P accurately encompassing the true MSE of noise across the majority of tasks.
This demonstrated their ability to accurately delineate the admissible regions for a given dataset.
KNIFE-P served as a practical upper bound for MSE, setting realistic expectations for model performance, while LMC-P established a strict lower bound, emphasizing the limitations of the predictability.
The results qualify KNIFE-P and LMC-P as appropriate for our application, highlighting their robustness and reliability.

\section{Application}
In the following context, we demonstrate how to use KNIFE-P and LMC-P for predictability analysis.
While MSE and MAE are relative indicators of performance, we extend the relationship between conditional entropy and MSE to the coefficient of determination \( R^2 \).
This extension from relative indicators to \( R^2 \) provides a more intuitive comprehension of the achievable predictability.

\subsection{Coefficient of Determination}
Among all the indicators of goodness-of-fit, we introduce the coefficient of determination $R^2$ into our application due to its relationship with MSE.
The coefficient of determination is
\[
  R^2(y)=1-\frac{\sum_{i=1}^{N}(y_i-\hat{y})^2}{\sum_{i=1}^{N}(y_i-\bar{y})^2}.
\]
Since we are deriving bounds for $R^2$, we assume noise $\epsilon$ following Gaussian distribution.
Within the three considered noise distributions, the Gaussian distribution has the lowest entropy for MSE.
The MSE of $y$ is bounded by
\[
  \frac{\sum_{i=1}^{N}(y_i-\hat{y})^2}{N}\geq \frac{1}{12}e^{2H(Y|X)}.
\]
One can see that the resemblance of coefficient of determination and MSE.
Hence, we can derive
\[
  R^2(y)=1-\frac{\sum_{i=1}^{N}(y_i-\hat{y})^2}{\sum_{i=1}^{N}(y_i-\bar{y})^2}\leq 1-\frac{Ne^{2H(Y|X)}}{12\sum_{i=1}^{N}(y_i-\bar{y})^2} =1-\frac{e^{2H(Y|X)}}{12\sigma_y} .
\]
Conditional entropy also bounds coefficient of determination.
As coefficient of determination being a indicator of goodness-of-fit, the extension from MSE to coefficient of determination enhances the interpretability of predictability analysis.

Assuming noise following a Gaussian distribution is practical for a bound of performance but conservative in specific conditions.
\citep{BF13} mentioned that long-tailed noise has a larger entropy compared to Gaussian noise.
The assumption of Gaussian noise can make optimistic suggestions.
Hence, for a more practical analysis, we refer not only to the achievable $R^2$ shown by estimators but also catch information from their differences.

We derive indicators with estimations $E\in\{knifecp, knifedp, lmcp\}$
\[
  R^2_{E}=1-\frac{e^{2\hat{H}^{E}}}{12\sigma_y}
\]
$R^2_{knifecp}$ indicates an achievable performance of $R^2$ while $R^2_{lmcp}$ poses an upperbound of $R^2$.
The use of $R^2_{knifecp}$ and  $R^2_{lmcp}$ indicates the predictability of features for the prediction of target $y$.
Generally, we should have the following relationship between these three indicators.
\[
  R^2_{knifedp} \leq R^2_{knifecp} \leq R^2_{lmcp} 
\]
As Theorem~\ref{Theorem:GeneralGap} and Theorem~\ref{Theorem:OptGap} suggested, the close values of $R^2_{knifecp}$, $R^2_{knifedp}$, and $R^2_{lmcp}$ indicate the lack of contribution of features in prediction.

\subsection{Experiment Setup}
Following an extensive study of \cite{MFD19}, we select datasets from the UCI Machine Learning Repository.
\begin{table}[!ht]
  \centering
  \caption{Selected datasets from the UCI repository \label{table:dataset}}
  \begin{tabular}{llrrl}
  \hline
  Original UCI name&Datasets & $\sigma^2_y$ & $N$ & Complexity \\ \hline
  Appliances energy prediction & app-energy & 10510.821 & 19735 & Difficult  \\
  Beijing PM2.5&BJ-pm25 & 8473.071 & 41757 & Difficult \\
  Bikes sharing&bike-hour & 32899.568 & 17379 & Easy  \\
  Relative location of CT slices on axial axis &CT-slices & 499.381 & 53500 & Difficult  \\
  Concrete compressive strength &comp-stren & 278.811 & 1030 & Easy  \\
   \hline
  \end{tabular}
\end{table}
The table lists the selected datasets from the UCI repository, detailing their original names, dataset names used in our study, the variance of the target variable $\sigma^2_y$, the number of instances $N$, and their complexity levels from \cite{MFD19}.

These datasets are selected due to their principal components.
We use principal component analysis (PCA) to decompose the variance of features.
By examining different levels of explained variance (EV), we create the degradation in predictability of features.
\begin{table}[h]
  \centering
  \caption{Principle complement analysis and parameter of selected datatset\label{table:pca}}
  \begin{tabular}{lrrrr}
  \hline
      Dataset & EV & $d$ & Hidden Dimension\\ \hline
      app-energy & 0.906 & 6 & 16 \\ 
      app-energy & 0.701 & 2 & 8  \\ 
      app-energy & 0.518 & 1 & 4  \\
      bike-hour & 0.919 & 3 & 8  \\ 
      bike-hour & 0.728 & 2 & 4  \\
      BJ-pm25 & 0.936 & 2 & 8  \\ 
      BJ-pm25 & 0.811 & 1 & 4  \\
      CT-slices & 0.901 & 121 & 256  \\ 
      CT-slices & 0.702 & 28 & 64   \\
      CT-slices & 0.516 & 7 & 32   \\ 
      CT-slices & 0.320 & 2 & 16   \\
      comp-stren & 0.969 & 5 & 8   \\
      comp-stren & 0.884 & 4 & 8  \\ 
      comp-stren & 0.774 & 3 & 8   \\
      comp-stren & 0.587 & 2 & 4   \\ 
      comp-stren & 0.333 & 1 & 4  \\ 
      \hline
  \end{tabular}
\end{table}
Table~\ref{table:pca} presents the principal component analysis (PCA) and parameters of the selected datasets.
For each task, we have the explained variance (EV) of components, the number of features $d$ (components), and the hidden dimension used in our estimation.
The datasets are separated into 60\% for training and 40\% for testing.

In \cite{MFD19}, tree models and gradient boosting based models are reported with outstaneding performance.
We select a few models as benchmarks.
M5 is a classical tree-based model \citep{JQ92}.
The parameters tuned are the flags for pruning and smoothing (each with values yes/no), and whether to create a tree or a rule set.

Another tree-based model selected is ExtraTree.
ExtraTree is an ensemble of extremely randomized regression trees \citep{PG06}.
It randomizes the input and cut-point of each split (or node in the tree) using a parameter that tunes the randomization strength.
The full training set is used instead of a bootstrap replica.
Explicit randomization of input and cut-point splittings combined with ensemble averaging is expected to reduce variance more effectively than other methods.
Its hyperparameters include the number of trees (ranging from 100 to 250) and the minimum sample size to split a node, tuned with integer values from 1 to 10.

Then, we select two frequently used gradient boosting machines \citep{JHF01}.
The first model, xgbTree, is the extreme gradient boosting model using trees as boosters \citep{TC16}.
The hyperparameters for xgbTree include maximum tree depth with values 3, 5, 7, 9, and 11, number of estimators ranging from 100 to 250, learning rate with values 0.3 and 0.4.
Lastly, we use another famous implementation, LightGBM \citep{GK17}.
We use two different settings for LightGBM: lgbforest uses random forest as the booster, while lgbtree uses trees. The hyperparameters for LightGBM include maximum tree depth (3, 5, 7, 9, and 11) and number of estimators (from 40 to 240).

\subsection{Dataset with Missing Information}
Table~\ref{table:reg} presents the performance of various models in terms of the validation $R^2$ for different datasets and EVs.
\begin{table}[!ht]
  \centering
  \caption{The $R^2$ of regression models \label{table:reg}}
  \begin{tabular}{lr|rrrrr}
    \hline
        Task & EV & lgbforest & lgbtree & m5 & extratree & xgboost  \\ \hline
        app-energy & 0.906 & 0.117 & 0.252 & 0.157 & \textbf{0.358} & 0.236  \\ 
        app-energy & 0.701 & \textbf{0.026} & -0.015 & -0.018 & -0.212 & -0.425  \\ 
        app-energy & 0.518 & \textbf{0.025} & 0.009 & 0.000 & -0.615 & 0.005  \\ 
        BJ-pm25 & 0.936 & \textbf{0.163} & \textbf{0.163} & -0.397 & 0.015 & -0.145  \\ 
        BJ-pm25 & 0.811 & \textbf{0.106} & 0.097 & -0.427 & -0.477 & 0.097  \\ 
        bike-hour & 0.919 & 0.601 & 0.654 & 0.576 & \textbf{0.679} & 0.648  \\ 
        bike-hour & 0.728 & 0.538 & \textbf{0.564} & 0.358 & 0.331 & \textbf{0.564}  \\ 
        CT-slices & 0.901 & 0.941 & 0.990 & 0.977 & \textbf{0.997} & 0.991  \\ 
        CT-slices & 0.702 & 0.940 & 0.989 & 0.973 & \textbf{0.998} & 0.991  \\ 
        CT-slices & 0.516 & 0.926 & 0.977 & 0.964 & \textbf{0.994} & 0.984  \\ 
        CT-slices & 0.320 & \textbf{0.498} & 0.541 & 0.525 & 0.482 & 0.403  \\ 
        comp-stren & 0.969 & 0.496 & \textbf{0.793} & 0.644 & 0.770 & 0.726  \\ 
        comp-stren & 0.884 & 0.448 & 0.601 & 0.456 & \textbf{0.673} & 0.598  \\ 
        comp-stren & 0.774 & 0.378 & 0.461 & 0.405 & \textbf{0.642} & 0.463  \\ 
        comp-stren & 0.587 & 0.318 & 0.420 & 0.292 & \textbf{0.537} & 0.357  \\ 
        comp-stren & 0.333 & 0.163 & 0.134 & \textbf{0.220} & -0.025 & 0.074 \\ \hline
    \end{tabular}
\end{table}
The models with the best \( R^2 \) values are highlighted in bold.
Table~\ref{table:indicator} shows estimated results of proposed \(R^2\) indicators. 
\begin{table}[!ht]
    \centering
    \caption{The estimated $R^2$ indicators \label{table:indicator}}
    \begin{tabular}{lr|rrr}
    \hline
        Task & EV & $R^2_{knifecp}$ & $R^2_{knifedp}$  & $R^2_{lmcp}$ \\ \hline
        app-energy & 0.906 & \textbf{0.789} & 0.746  & \textbf{0.832} \\ 
        app-energy & 0.701 & \textbf{0.777} & 0.758  & \textbf{0.779} \\
        app-energy & 0.518 & \textbf{0.760} & 0.747  & \textbf{0.767} \\ 
        BJ-pm25 & 0.936 & \textbf{0.469} & 0.316  & \textbf{0.580} \\
        BJ-pm25 & 0.811 & \textbf{0.401} & 0.309  & \textbf{0.475} \\ 
        bike-hour & 0.919 & \textbf{0.779} & 0.442  & \textbf{0.912} \\ 
        bike-hour & 0.728 & \textbf{0.730} & 0.423  & \textbf{0.869} \\ 
        CT-slices & 0.901 & 0.810 & 0.128  & 0.991 \\
        CT-slices & 0.702 & 0.930 & 0.131  & \textbf{1.000} \\
        CT-slices & 0.516 & 0.957 & 0.128  & \textbf{1.000} \\
        CT-slices & 0.320 & \textbf{0.863} & 0.130  & \textbf{0.992} \\
        comp-stren & 0.969 & 0.616 & 0.038  & \textbf{0.812} \\
        comp-stren & 0.884 & 0.479 & 0.027  & \textbf{0.718} \\
        comp-stren & 0.774 & 0.369 & 0.055  & \textbf{0.651} \\ 
        comp-stren & 0.587 & 0.315 & 0.043  & \textbf{0.539} \\
        comp-stren & 0.333 & 0.220 & 0.052 & \textbf{0.356} \\ \hline
    \end{tabular}
\end{table}
If the $R^2_{knifecp}$ or $R^2_{lmcp}$ successfully bounds the best \( R^2 \) among regression models, their values are highlighted in bold.
Variance of models and indicators are presented in \ref{append-pra-var}.

Most estimators effectively responded to decreasing EV levels, with $R^2_{knifecp}$ and $R^2_{lmcp}$ bounding the best $R^2$ across the majority of tasks.
This demonstrated their ability to delineate the achievable predictability for a given dataset.

For the app-energy dataset, the performance of the models is not ideal. 
he values of \( R^2_{knifecp} \) and \( R^2_{knifedp} \) are close, indicating that the features lack sufficient information to predict the target variable, especially with EV of 0.701 and 0.518. 
Furthermore, the \( R^2_{knifedp} \) value of around 0.75 suggests that the distribution of the target variable may be approximated without contributions from the features.

For the BJ-pm25 dataset, the lgbforest and lgbtree models tie with the highest \( R^2 \) value of 0.163 at an EV of 0.936, while lgbforest achieves another high \( R^2 \) value of 0.106 at an EV of 0.811.
The values of \( R^2_{knifecp} \), \( R^2_{knifedp} \), and \( R^2_{lmcp} \) are not close, suggesting that the prediction can be mode from the features.
However, their low values also imply limitations in predicting the target variable for BJ-pm25.
\( R^2_{knifecp} \) provides achievable \( R^2 \) values ranging from 0.401 to 0.469, while \( R^2_{lmcp} \) upper bounds the achievable $R^2$ from 0.475 to 0.580.

The bike-hour dataset serves as a classical demonstration for our indicators.
With an EV of 0.919, the extratree model leads with an \( R^2 \) value of 0.679.
Following that, xgboost and lgbtree tie at an EV of 0.728, both achieving an \( R^2 \) value of 0.564. 
\( R^2_{knifedp} \) has low values below 0.45, while \( R^2_{knifecp} \) has values over 0.72 indicating the potential of the features to predict the target variable.
Additionally, \( R^2_{lmcp} \) upper bounds are between 0.869 and 0.912, further implies the existence of model with high predictability.

For the CT-slices dataset with an EV of 0.901, the extratree model achieves the highest \( R^2 \) value of 0.997, surpassing other models such as lgbtree (0.990) and xgbTree (0.991).
As the EV decreases to 0.702 and 0.516, the extratree model consistently maintains the highest \( R^2 \) values of 0.998 and 0.994, respectively.
At an EV of 0.320, lgbtree takes the lead with an \( R^2 \) value of 0.541.
The large difference between \( R^2_{knifecp} \) and \( R^2_{knifedp} \) suggests the features' ability to predict the target variable.
As a result, the regression models generally perform well on this task.

However, as the EV drops to 0.320, the models' performance also declines.
Correspondingly, \( R^2_{knifecp} \) decreases in response to the reduced predictability.
One drawback of our indicator is shown in this task: with an EV of 0.901, \( R^2_{lmcp} \) fails to bound the best \( R^2 \).
This may be due to the curse of dimensionality affecting the fitting of the estimator.

For the comp-stren dataset with an EV of 0.969, lgbtree achieves the highest \( R^2 \) value of 0.793, while extratree performs best for EV values of 0.884, 0.774, and 0.587 with corresponding \( R^2 \) values.
The m5 model takes the lead with an EV of 0.333.
This case demonstrates the sensitivity of our indicators.
At an EV of 0.969, the large difference between \( R^2_{knifecp} \) and \( R^2_{knifedp} \) indicates the good predictability of the given features and target variables.
As the EV decreases, \( R^2_{knifecp} \) drops accordingly, reflecting the decline in predictability.

In summary, our proposed indicators \( R^2_{knifecp} \), \( R^2_{knifedp} \), and \( R^2_{lmcp} \) proved to be reliable tools for evaluating the predictability of regression models.
While indicators suggests low $R^2$ or closed performance between \( R^2_{knifecp} \) and \( R^2_{knifedp} \), the regression models affirm the suggestion of the proposed indicators.
The proposed indicators effectively captured the nuances of feature-target relationships across diverse datasets, providing valuable insights into model performance and the inherent predictability of the data.
\( R^2_{lmcp} \) only fails bound the best performance in one task with 121 features.
Despite some limitations, related to the curse of dimensionality, proposed indicators provides a robust predictability analysis with a direct inference to model performance.

\section{Conclusion}
This study introduces predictability analysis into regression problems in machine learning.
While traditional error metrics like MSE and MAE offer insights into model accuracy, they cannot separate the design of model and the contribution of features.
By introducing conditional entropy as an estimator for predictability, this research bridges the gap in understanding feature contribution in regression problems.

Learned from the previous estimators KNIFE and CLUB, the developed conditional entropy KNIFE-P estimator and LMC-P estimator offer reliable predictions by providing both lower and upper bounds for conditional entropy.
These bounds enable a practical framework for evaluating predictability beyond conventional error metrics.
The robustness of these estimators was demonstrated through experiments on synthesized datasets and real-world datasets from the UCI Machine Learning Repository, showcasing their applicability and relevance.

The proposed approach extends the analysis to the coefficient of determination \( R^2 \), enhancing the interpretability of predictability analysis.
The results indicate that KNIFE-P and LMC-P effectively capture the admissible region for prediction models, delineating the achievable performance and the limitations of feature sets.
Despite some limitations related to the curse of dimensionality, these indicators provide a robust analysis framework for assessing the predictability of regression models.

Overall, this study contributes to a deeper understanding of achievable model performance and feature information in making predictions, providing valuable tools for assessing feature relevance.
Proposed indicators aid in the development of regression problems by providing a predictability analysis with a direct inference to model performance.
Also, it can be applied to classification problems with subtle adaptations.
Future research can explore further enhancements, such as integration with long-tail tests, a more efficient estimator, or a more light weighted estimator.
\bibliography{CondEntropy}

\begin{thebibliography}{37}
\providecommand{\natexlab}[1]{#1}
\providecommand{\url}[1]{\texttt{#1}}
\expandafter\ifx\csname urlstyle\endcsname\relax
  \providecommand{\doi}[1]{doi: #1}\else
  \providecommand{\doi}{doi: \begingroup \urlstyle{rm}\Url}\fi

\bibitem[Battiti(1994)]{RB94}
Roberto Battiti.
\newblock Using mutual information for selecting features in supervised neural net learning.
\newblock \emph{IEEE Transactions on neural networks}, 5\penalty0 (4):\penalty0 537--550, 1994.

\bibitem[Belghazi et~al.(2018)Belghazi, Baratin, Rajeshwar, Ozair, Bengio, Courville, and Hjelm]{MIB18}
Mohamed~Ishmael Belghazi, Aristide Baratin, Sai Rajeshwar, Sherjil Ozair, Yoshua Bengio, Aaron Courville, and Devon Hjelm.
\newblock Mutual {Information} {Neural} {Estimation}.
\newblock In \emph{Proceedings of the 35th {International} {Conference} on {Machine} {Learning}}, pages 531--540. Proceedings of Machine Learning Research, July 2018.
\newblock URL \url{https://proceedings.mlr.press/v80/belghazi18a.html}.
\newblock ISSN: 2640-3498.

\bibitem[Chen and Guestrin(2016)]{TC16}
Tianqi Chen and Carlos Guestrin.
\newblock Xgboost: A scalable tree boosting system.
\newblock In \emph{Proceedings of the 22nd acm sigkdd international conference on knowledge discovery and data mining}, pages 785--794, 2016.

\bibitem[Cheng et~al.(2020)Cheng, Hao, Dai, Liu, Gan, and Carin]{PC20}
Pengyu Cheng, Weituo Hao, Shuyang Dai, Jiachang Liu, Zhe Gan, and Lawrence Carin.
\newblock Club: A contrastive log-ratio upper bound of mutual information.
\newblock In \emph{International conference on machine learning}, pages 1779--1788. Proceedings of Machine Learning Research, 2020.

\bibitem[Degeest et~al.(2021)Degeest, Frénay, and Verleysen]{DFV21}
Alexandra Degeest, Benoît Frénay, and Michel Verleysen.
\newblock Reading grid for feature selection relevance criteria in regression.
\newblock \emph{Pattern Recognition Letters}, 148:\penalty0 92--99, August 2021.
\newblock ISSN 0167-8655.
\newblock \doi{10.1016/j.patrec.2021.04.031}.
\newblock URL \url{https://www.sciencedirect.com/science/article/pii/S0167865521001768}.

\bibitem[Fern{\'a}ndez-Delgado et~al.(2019)Fern{\'a}ndez-Delgado, Sirsat, Cernadas, Alawadi, Barro, and Febrero-Bande]{MFD19}
Manuel Fern{\'a}ndez-Delgado, Manisha~Sanjay Sirsat, Eva Cernadas, Sadi Alawadi, Sen{\'e}n Barro, and Manuel Febrero-Bande.
\newblock An extensive experimental survey of regression methods.
\newblock \emph{Neural Networks}, 111:\penalty0 11--34, 2019.

\bibitem[Friedman(2001)]{JHF01}
Jerome~H Friedman.
\newblock Greedy function approximation: a gradient boosting machine.
\newblock \emph{Annals of statistics}, pages 1189--1232, 2001.

\bibitem[Frénay et~al.(2013)Frénay, Doquire, and Verleysen]{BF13}
Benoît Frénay, Gauthier Doquire, and Michel Verleysen.
\newblock Is mutual information adequate for feature selection in regression?
\newblock \emph{Neural Networks}, 48:\penalty0 1--7, December 2013.
\newblock ISSN 0893-6080.
\newblock \doi{10.1016/j.neunet.2013.07.003}.
\newblock URL \url{https://www.sciencedirect.com/science/article/pii/S0893608013001883}.

\bibitem[Geurts et~al.(2006)Geurts, Ernst, and Wehenkel]{PG06}
Pierre Geurts, Damien Ernst, and Louis Wehenkel.
\newblock Extremely randomized trees.
\newblock \emph{Machine learning}, 63:\penalty0 3--42, 2006.

\bibitem[Haven et~al.(2005)Haven, Majda, and Abramov]{KH05}
Kyle Haven, Andrew Majda, and Rafail Abramov.
\newblock Quantifying predictability through information theory: small sample estimation in a non-gaussian framework.
\newblock \emph{Journal of Computational Physics}, 206\penalty0 (1):\penalty0 334--362, 2005.

\bibitem[Huber et~al.(2008)Huber, Bailey, Durrant-Whyte, and Hanebeck]{MFH08}
Marco~F Huber, Tim Bailey, Hugh Durrant-Whyte, and Uwe~D Hanebeck.
\newblock On entropy approximation for gaussian mixture random vectors.
\newblock In \emph{2008 IEEE International Conference on Multisensor Fusion and Integration for Intelligent Systems}, pages 181--188. IEEE, 2008.

\bibitem[Ke et~al.(2017)Ke, Meng, Finley, Wang, Chen, Ma, Ye, and Liu]{GK17}
Guolin Ke, Qi~Meng, Thomas Finley, Taifeng Wang, Wei Chen, Weidong Ma, Qiwei Ye, and Tie-Yan Liu.
\newblock Lightgbm: A highly efficient gradient boosting decision tree.
\newblock \emph{Advances in neural information processing systems}, 30:\penalty0 3146--3154, 2017.

\bibitem[Krishnamurthy(2019)]{VK19}
Venkataramanaiah Krishnamurthy.
\newblock Predictability of weather and climate.
\newblock \emph{Earth and Space Science}, 6\penalty0 (7):\penalty0 1043--1056, 2019.

\bibitem[Kullback(1997)]{SK97}
Solomon Kullback.
\newblock \emph{Information theory and statistics}.
\newblock Courier Corporation, 1997.

\bibitem[Li et~al.(2022)Li, Knoop, and van Lint]{GL22}
Guopeng Li, Victor~L Knoop, and Hans van Lint.
\newblock Estimate the limit of predictability in short-term traffic forecasting: An entropy-based approach.
\newblock \emph{Transportation Research Part C: Emerging Technologies}, 138:\penalty0 103607, 2022.

\bibitem[Li et~al.(2017)Li, Cheng, Wang, Morstatter, Trevino, Tang, and Liu]{JL17}
Jundong Li, Kewei Cheng, Suhang Wang, Fred Morstatter, Robert~P Trevino, Jiliang Tang, and Huan Liu.
\newblock Feature selection: A data perspective.
\newblock \emph{ACM computing surveys (CSUR)}, 50\penalty0 (6):\penalty0 1--45, 2017.

\bibitem[Lu et~al.(2017)Lu, Pu, Wang, Hu, and Wang]{ZL17}
Zhou Lu, Hongming Pu, Feicheng Wang, Zhiqiang Hu, and Liwei Wang.
\newblock The expressive power of neural networks: A view from the width.
\newblock \emph{Advances in neural information processing systems}, 30, 2017.

\bibitem[Maz'ya and Schmidt(1996)]{VM96}
Vladimir Maz'ya and Gunther Schmidt.
\newblock On approximate approximations using gaussian kernels.
\newblock \emph{IMA Journal of Numerical Analysis}, 16\penalty0 (1):\penalty0 13--29, 1996.

\bibitem[McAllester and Stratos(2020)]{DM20}
David McAllester and Karl Stratos.
\newblock Formal limitations on the measurement of mutual information.
\newblock In \emph{International Conference on Artificial Intelligence and Statistics}, pages 875--884. Proceedings of Machine Learning Research, 2020.

\bibitem[Nair et~al.(2001)Nair, Liu, Rilett, and Gupta]{AN01}
Attoor~Sanju Nair, Jyh-Charn Liu, Laurence Rilett, and Saurabh Gupta.
\newblock Non-linear analysis of traffic flow.
\newblock In \emph{ITSC 2001. 2001 IEEE Intelligent Transportation Systems. Proceedings (Cat. No. 01TH8585)}, pages 681--685. IEEE, 2001.

\bibitem[Nguyen et~al.(2010)Nguyen, Wainwright, and Jordan]{XLN10}
XuanLong Nguyen, Martin~J Wainwright, and Michael~I Jordan.
\newblock Estimating divergence functionals and the likelihood ratio by convex risk minimization.
\newblock \emph{IEEE Transactions on Information Theory}, 56\penalty0 (11):\penalty0 5847--5861, 2010.

\bibitem[Oord et~al.(2018)Oord, Li, and Vinyals]{AO18}
Aaron van~den Oord, Yazhe Li, and Oriol Vinyals.
\newblock Representation learning with contrastive predictive coding.
\newblock \emph{arXiv preprint arXiv:1807.03748}, 2018.

\bibitem[Oymak and Soltanolkotabi(2019)]{SO19}
Samet Oymak and Mahdi Soltanolkotabi.
\newblock Overparameterized nonlinear learning: Gradient descent takes the shortest path?
\newblock In \emph{International Conference on Machine Learning}, pages 4951--4960. PMLR, 2019.

\bibitem[Pedregosa et~al.(2011)Pedregosa, Varoquaux, Gramfort, Michel, Thirion, Grisel, Blondel, Prettenhofer, Weiss, Dubourg, Vanderplas, Passos, Cournapeau, Brucher, Perrot, and Duchesnay]{SL11}
F.~Pedregosa, G.~Varoquaux, A.~Gramfort, V.~Michel, B.~Thirion, O.~Grisel, M.~Blondel, P.~Prettenhofer, R.~Weiss, V.~Dubourg, J.~Vanderplas, A.~Passos, D.~Cournapeau, M.~Brucher, M.~Perrot, and E.~Duchesnay.
\newblock Scikit-learn: Machine learning in {P}ython.
\newblock \emph{Journal of Machine Learning Research}, 12:\penalty0 2825--2830, 2011.

\bibitem[Pichler et~al.(2022)Pichler, Colombo, Boudiaf, Koliander, and Piantanida]{GP22}
Georg Pichler, Pierre Jean~A Colombo, Malik Boudiaf, G{\"u}nther Koliander, and Pablo Piantanida.
\newblock A differential entropy estimator for training neural networks.
\newblock In \emph{International Conference on Machine Learning}, pages 17691--17715. Proceedings of Machine Learning Research, 2022.

\bibitem[Quinlan et~al.(1992)]{JQ92}
John~R Quinlan et~al.
\newblock Learning with continuous classes.
\newblock In \emph{5th Australian joint conference on artificial intelligence}, volume~92, pages 343--348. World Scientific, 1992.

\bibitem[Raghu et~al.(2017)Raghu, Poole, Kleinberg, Ganguli, and Sohl-Dickstein]{MR17}
Maithra Raghu, Ben Poole, Jon Kleinberg, Surya Ganguli, and Jascha Sohl-Dickstein.
\newblock On the expressive power of deep neural networks.
\newblock In \emph{international conference on machine learning}, pages 2847--2854. PMLR, 2017.

\bibitem[Rothfuss et~al.(2019)Rothfuss, Ferreira, Walther, and Ulrich]{JR19}
Jonas Rothfuss, Fabio Ferreira, Simon Walther, and Maxim Ulrich.
\newblock Conditional density estimation with neural networks: Best practices and benchmarks, 2019.

\bibitem[Rothfuss et~al.(2020)Rothfuss, Ferreira, Boehm, Walther, Ulrich, Asfour, and Krause]{JR20}
Jonas Rothfuss, Fabio Ferreira, Simon Boehm, Simon Walther, Maxim Ulrich, Tamim Asfour, and Andreas Krause.
\newblock Noise regularization for conditional density estimation, 2020.

\bibitem[Shang et~al.(2005)Shang, Li, and Kamae]{PS05}
Pengjian Shang, Xuewei Li, and Santi Kamae.
\newblock Chaotic analysis of traffic time series.
\newblock \emph{Chaos, Solitons \& Fractals}, 25\penalty0 (1):\penalty0 121--128, 2005.

\bibitem[Shannon(1948)]{CES48}
Claude~Elwood Shannon.
\newblock A mathematical theory of communication.
\newblock \emph{The Bell system technical journal}, 27\penalty0 (3):\penalty0 379--423, 1948.

\bibitem[Song et~al.(2010)Song, Qu, Blumm, and Barab{\'a}si]{CS10}
Chaoming Song, Zehui Qu, Nicholas Blumm, and Albert-L{\'a}szl{\'o} Barab{\'a}si.
\newblock Limits of predictability in human mobility.
\newblock \emph{Science}, 327\penalty0 (5968):\penalty0 1018--1021, 2010.

\bibitem[Stulp and Sigaud(2015)]{FS15}
Freek Stulp and Olivier Sigaud.
\newblock Many regression algorithms, one unified model: A review.
\newblock \emph{Neural Networks}, 69:\penalty0 60--79, 2015.

\bibitem[Takaguchi et~al.(2011)Takaguchi, Nakamura, Sato, Yano, and Masuda]{TT11}
Taro Takaguchi, Mitsuhiro Nakamura, Nobuo Sato, Kazuo Yano, and Naoki Masuda.
\newblock Predictability of conversation partners.
\newblock \emph{Physical Review X}, 1\penalty0 (1):\penalty0 011008, 2011.

\bibitem[Vergara and Est{\'e}vez(2014)]{JRV14}
Jorge~R Vergara and Pablo~A Est{\'e}vez.
\newblock A review of feature selection methods based on mutual information.
\newblock \emph{Neural computing and applications}, 24:\penalty0 175--186, 2014.

\bibitem[Wang et~al.(2020)Wang, Yan, and Ma]{XW20}
Xiujuan Wang, Yixuan Yan, and Xiaoyue Ma.
\newblock Feature selection method based on differential correlation information entropy.
\newblock \emph{Neural Processing Letters}, 52:\penalty0 1339--1358, 2020.

\bibitem[Wolf et~al.(1985)Wolf, Swift, Swinney, and Vastano]{AW85}
Alan Wolf, Jack~B Swift, Harry~L Swinney, and John~A Vastano.
\newblock Determining lyapunov exponents from a time series.
\newblock \emph{Physica D: nonlinear phenomena}, 16\penalty0 (3):\penalty0 285--317, 1985.

\end{thebibliography}
\begin{appendices}
\section{Variance of Experiments}
\subsection{Variance of Estimators in Synthesized Experiment\label{append-syn-var}}
Here we present the estimation variance of the estimators in the synthesized experiment.
Variance can be found in Table~\ref{table:knifep-syn-var}, Table~\ref{table:lmc-syn-var}, Table~\ref{table:mine-syn-var}, Table~\ref{table:nce-syn-var}, and Table~\ref{table:knife-gauss-syn-var}.
Notice that these values are not converted by relationship in Table~\ref{table:relationship}.
The target indicates conditional entropy as CE, differential entropy as DE, and mutual information as MI.
The column ND is the noise distribution: 0 for uniform noise, 1 for laplacian noise, and 2 for normal noise.
The tasks are ordered as follows: 0 for linearity, 1 for nonlinearity, 2 for interaction, and 3 for multivariate.
\begin{table}[]
  \centering
  \caption{Variance of KNIFE-P in synthesized experiment~\label{table:knifep-syn-var}}
  \begin{tabular}{llrrrrrrr}
  \hline
  Estimator & target & ND & task & 10       & 250      & 500      & 750      & 1000     \\
  \hline
  KNIFE-P   & CE     & 0  & 0    & 8.79E-05 & 1.09E-04 & 7.35E-05 & 1.46E-04 & 6.30E-05 \\
  KNIFE-P   & CE     & 1  & 0    & 1.87E-04 & 1.10E-03 & 1.06E-03 & 8.01E-04 & 2.02E-03 \\
  KNIFE-P   & CE     & 2  & 0    & 1.70E-04 & 3.80E-04 & 5.60E-04 & 1.36E-03 & 3.28E-04 \\
  KNIFE-P   & CE     & 0  & 1    & 1.85E-04 & 5.67E-04 & 5.44E-05 & 1.13E-04 & 1.50E-04 \\
  KNIFE-P   & CE     & 1  & 1    & 9.47E-05 & 4.13E-04 & 8.53E-05 & 1.37E-03 & 7.97E-04 \\
  KNIFE-P   & CE     & 2  & 1    & 1.52E-04 & 1.86E-04 & 5.79E-04 & 5.51E-04 & 3.95E-04 \\
  KNIFE-P   & CE     & 0  & 2    & 2.15E-02 & 9.11E-04 & 1.00E-01 & 1.06E-01 & 1.65E-02 \\
  KNIFE-P   & CE     & 1  & 2    & 1.50E-02 & 4.56E-03 & 5.41E-03 & 5.50E-04 & 7.01E-04 \\
  KNIFE-P   & CE     & 2  & 2    & 3.47E-03 & 1.11E-02 & 4.40E-04 & 2.05E-02 & 2.54E-03 \\
  KNIFE-P   & CE     & 0  & 3    & 4.22E-03 & 5.35E-04 & 4.77E-04 & 6.98E-05 & 1.41E-04 \\
  KNIFE-P   & CE     & 1  & 3    & 3.10E-04 & 5.49E-02 & 2.87E-03 & 3.55E-04 & 4.36E-03 \\
  KNIFE-P   & CE     & 2  & 3    & 1.32E-03 & 7.12E-04 & 3.21E-04 & 1.11E-03 & 2.94E-04 \\
  KNIFE-P   & DE     & 0  & 0    & 7.90E-05 & 3.98E-04 & 1.87E-04 & 3.92E-04 & 3.98E-04 \\
  KNIFE-P   & DE     & 1  & 0    & 1.08E-04 & 2.21E-04 & 1.15E-04 & 7.00E-04 & 1.43E-03 \\
  KNIFE-P   & DE     & 2  & 0    & 3.02E-05 & 3.00E-04 & 5.17E-04 & 7.23E-04 & 2.77E-04 \\
  KNIFE-P   & DE     & 0  & 1    & 4.84E-05 & 6.68E-05 & 2.95E-04 & 1.17E-04 & 2.10E-04 \\
  KNIFE-P   & DE     & 1  & 1    & 2.45E-05 & 7.53E-06 & 2.69E-04 & 1.52E-03 & 9.07E-04 \\
  KNIFE-P   & DE     & 2  & 1    & 9.56E-05 & 2.13E-04 & 6.42E-04 & 6.14E-04 & 5.05E-04 \\
  KNIFE-P   & DE     & 0  & 2    & 4.63E-04 & 6.13E-04 & 1.79E-03 & 1.23E-04 & 1.36E-04 \\
  KNIFE-P   & DE     & 1  & 2    & 7.53E-04 & 2.01E-03 & 4.46E-04 & 2.33E-04 & 4.74E-04 \\
  KNIFE-P   & DE     & 2  & 2    & 2.59E-04 & 1.12E-03 & 3.83E-04 & 8.55E-04 & 2.47E-03 \\
  KNIFE-P   & DE     & 0  & 3    & 1.92E-04 & 3.48E-04 & 3.55E-04 & 1.33E-04 & 9.78E-05 \\
  KNIFE-P   & DE     & 1  & 3    & 2.89E-04 & 6.70E-04 & 2.24E-03 & 1.55E-04 & 4.31E-03 \\
  KNIFE-P   & DE     & 2  & 3    & 1.63E-04 & 2.36E-04 & 6.34E-04 & 5.88E-04 & 7.14E-04 \\
  \hline
  \end{tabular}
  \end{table}
\begin{table}[]
  \centering
  \caption{Variance of LMC-P estimator in synthesized experiment~\label{table:lmc-syn-var}}
  \begin{tabular}{llrrrrrrr}
  \hline
Estimator & target & ND & task & 10       & 250      & 500      & 750      & 1000     \\
\hline  
LMC-P     & CE     & 0  & 0    & 2.80E-02 & 1.52E-02 & 2.15E-03 & 7.03E-03 & 3.94E-03 \\
LMC-P     & CE     & 1  & 0    & 1.93E-02 & 1.03E-02 & 4.87E-04 & 1.02E-03 & 1.64E-03 \\
LMC-P     & CE     & 2  & 0    & 3.14E-02 & 7.88E-03 & 3.42E-03 & 3.43E-03 & 1.25E-03 \\
LMC-P     & CE     & 0  & 1    & 3.36E-03 & 1.14E-02 & 4.86E-03 & 2.72E-03 & 4.27E-04 \\
LMC-P     & CE     & 1  & 1    & 1.19E-02 & 7.67E-03 & 9.95E-04 & 2.96E-03 & 4.04E-04 \\
LMC-P     & CE     & 2  & 1    & 5.03E-04 & 1.42E-03 & 7.94E-03 & 2.42E-03 & 4.38E-03 \\
LMC-P     & CE     & 0  & 2    & 6.38E-02 & 3.41E-03 & 4.42E-03 & 4.36E-03 & 2.91E-03 \\
LMC-P     & CE     & 1  & 2    & 5.45E-02 & 1.39E-02 & 3.04E-03 & 4.78E-04 & 1.51E-03 \\
LMC-P     & CE     & 2  & 2    & 2.65E-02 & 1.82E-02 & 1.51E-03 & 3.23E-03 & 3.53E-03 \\
LMC-P     & CE     & 0  & 3    & 1.45E-01 & 5.98E-03 & 8.20E-04 & 9.06E-04 & 1.90E-03 \\
LMC-P     & CE     & 1  & 3    & 9.71E-02 & 4.02E-02 & 2.29E-03 & 3.47E-03 & 1.65E-03 \\
LMC-P     & CE     & 2  & 3    & 4.74E-02 & 1.45E-02 & 5.29E-03 & 2.44E-03 & 2.39E-03 \\
\hline  
\end{tabular}
  \end{table}

\begin{table}[]
  \centering
  \caption{Variance of MINE estimator in synthesized experiment~\label{table:mine-syn-var}}
  \begin{tabular}{llrrrrrrr}
    \hline
  Estimator & target & ND & task & 10       & 250      & 500      & 750      & 1000     \\
  \hline
  MINE      & DE     & 0  & 0    & 1.87E-01 & 6.34E-01 & 4.65E-01 & 4.39E-01 & 3.51E-01 \\
  MINE      & DE     & 1  & 0    & 2.62E-01 & 1.87E+00 & 6.66E-01 & 2.10E+00 & 5.06E-01 \\
  MINE      & DE     & 2  & 0    & 2.99E-01 & 3.93E+00 & 1.29E+00 & 1.09E+00 & 1.57E+00 \\
  MINE      & DE     & 0  & 1    & 1.96E-01 & 1.26E+00 & 1.89E+00 & 1.60E+00 & 3.57E+00 \\
  MINE      & DE     & 1  & 1    & 1.06E+00 & 1.05E+00 & 1.79E+00 & 7.76E-01 & 6.56E-01 \\
  MINE      & DE     & 2  & 1    & 3.83E-01 & 2.05E-01 & 3.94E+00 & 6.84E-01 & 7.39E-01 \\
  MINE      & DE     & 0  & 2    & 6.45E-02 & 3.69E-01 & 1.85E+00 & 5.30E-01 & 1.42E-01 \\
  MINE      & DE     & 1  & 2    & 9.10E-01 & 6.88E-01 & 2.08E+00 & 7.37E-02 & 2.16E+00 \\
  MINE      & DE     & 2  & 2    & 3.19E+00 & 1.78E+00 & 1.55E+00 & 2.47E+00 & 6.46E-01 \\
  MINE      & DE     & 0  & 3    & 4.66E-01 & 3.83E+00 & 5.85E-01 & 3.73E-01 & 3.25E-01 \\
  MINE      & DE     & 1  & 3    & 2.02E-01 & 3.13E-01 & 1.71E-01 & 1.93E-01 & 3.69E-01 \\
  MINE      & DE     & 2  & 3    & 1.17E+00 & 3.83E+00 & 5.60E-01 & 7.87E-01 & 9.70E-01 \\
  MINE      & MI     & 0  & 0    & 1.52E-04 & 1.47E-04 & 1.48E-04 & 1.02E-04 & 5.50E-04 \\
  MINE      & MI     & 1  & 0    & 2.08E-04 & 2.99E-04 & 1.09E-04 & 1.38E-04 & 3.66E-04 \\
  MINE      & MI     & 2  & 0    & 6.34E-04 & 9.55E-05 & 1.46E-04 & 5.01E-05 & 8.14E-05 \\
  MINE      & MI     & 0  & 1    & 8.33E-04 & 1.97E-04 & 4.59E-04 & 1.47E-04 & 7.92E-05 \\
  MINE      & MI     & 1  & 1    & 1.47E-03 & 1.52E-04 & 1.43E-04 & 2.23E-04 & 4.24E-04 \\
  MINE      & MI     & 2  & 1    & 2.20E-03 & 1.35E-04 & 6.58E-05 & 1.43E-04 & 1.49E-04 \\
  MINE      & MI     & 0  & 2    & 2.24E-04 & 1.33E-04 & 3.18E-04 & 4.21E-04 & 1.75E-05 \\
  MINE      & MI     & 1  & 2    & 4.36E-04 & 3.97E-04 & 1.14E-03 & 4.54E-04 & 2.62E-04 \\
  MINE      & MI     & 2  & 2    & 2.47E-04 & 2.31E-04 & 8.91E-04 & 7.69E-04 & 1.23E-04 \\
  MINE      & MI     & 0  & 3    & 1.27E-03 & 5.96E-04 & 1.01E-02 & 1.76E-02 & 1.93E-02 \\
  MINE      & MI     & 1  & 3    & 4.13E-03 & 1.06E-03 & 2.39E-03 & 2.25E-02 & 1.27E-02 \\
  MINE      & MI     & 2  & 3    & 1.05E-03 & 2.80E-03 & 1.02E-02 & 3.53E-03 & 2.68E-03 \\
  \hline
  \end{tabular}
  \end{table}
  \begin{table}[]
    \centering
    \caption{Variance of InfoNCE estimator in synthesized experiment~\label{table:nce-syn-var}}
    \begin{tabular}{llrrrrrrr}
    \hline
    Estimator & target & ND & task & 10       & 250      & 500      & 750      & 1000     \\
    \hline
    InfoNCE   & DE     & 0  & 0    & 5.74E-02 & 3.03E-01 & 8.28E-02 & 2.18E-01 & 1.50E-01 \\
    InfoNCE   & DE     & 1  & 0    & 2.96E+00 & 4.17E-02 & 7.15E-02 & 1.83E-01 & 3.67E-01 \\
    InfoNCE   & DE     & 2  & 0    & 3.65E-01 & 6.50E-01 & 5.65E-01 & 1.06E+00 & 6.22E-01 \\
    InfoNCE   & DE     & 0  & 1    & 5.25E-01 & 1.54E-01 & 2.77E-01 & 2.47E-02 & 1.18E-01 \\
    InfoNCE   & DE     & 1  & 1    & 2.67E-01 & 6.81E-02 & 3.51E-01 & 1.37E-01 & 7.02E-01 \\
    InfoNCE   & DE     & 2  & 1    & 7.76E-01 & 3.56E-02 & 1.86E-01 & 9.37E-01 & 2.06E-01 \\
    InfoNCE   & DE     & 0  & 2    & 4.55E-01 & 1.22E+00 & 1.49E-02 & 2.82E-01 & 2.82E-02 \\
    InfoNCE   & DE     & 1  & 2    & 2.67E+00 & 2.43E+00 & 5.19E-01 & 1.01E-02 & 9.21E-01 \\
    InfoNCE   & DE     & 2  & 2    & 9.84E-02 & 4.03E-02 & 6.28E+00 & 1.06E+00 & 8.41E-02 \\
    InfoNCE   & DE     & 0  & 3    & 8.54E-02 & 5.11E-01 & 7.82E-01 & 4.53E-01 & 2.53E-01 \\
    InfoNCE   & DE     & 1  & 3    & 3.39E-01 & 1.24E+00 & 2.20E-01 & 6.72E-02 & 4.56E-02 \\
    InfoNCE   & DE     & 2  & 3    & 2.56E-01 & 1.95E-02 & 1.22E-01 & 4.52E-01 & 8.03E-01 \\
    InfoNCE   & MI     & 0  & 0    & 2.77E-05 & 1.24E-04 & 1.33E-04 & 8.64E-05 & 1.04E-04 \\
    InfoNCE   & MI     & 1  & 0    & 2.31E-04 & 2.71E-04 & 1.98E-04 & 2.63E-05 & 3.18E-04 \\
    InfoNCE   & MI     & 2  & 0    & 2.19E-04 & 3.65E-04 & 1.42E-04 & 2.79E-04 & 1.20E-04 \\
    InfoNCE   & MI     & 0  & 1    & 2.50E-05 & 6.76E-04 & 7.98E-04 & 8.32E-04 & 2.13E-04 \\
    InfoNCE   & MI     & 1  & 1    & 4.99E-05 & 2.18E-04 & 5.97E-05 & 2.91E-04 & 4.15E-04 \\
    InfoNCE   & MI     & 2  & 1    & 4.38E-04 & 7.35E-05 & 1.66E-04 & 2.45E-04 & 3.89E-04 \\
    InfoNCE   & MI     & 0  & 2    & 1.18E-03 & 2.55E-02 & 2.17E-02 & 2.42E-04 & 4.64E-04 \\
    InfoNCE   & MI     & 1  & 2    & 4.93E-03 & 8.25E-03 & 2.31E-02 & 1.56E-02 & 6.13E-03 \\
    InfoNCE   & MI     & 2  & 2    & 1.94E-03 & 2.46E-02 & 2.13E-03 & 4.76E-03 & 1.98E-03 \\
    InfoNCE   & MI     & 0  & 3    & 1.97E-03 & 3.34E-03 & 1.77E-02 & 2.33E-02 & 7.40E-03 \\
    InfoNCE   & MI     & 1  & 3    & 1.23E-03 & 9.64E-03 & 6.00E-03 & 4.56E-03 & 1.29E-02 \\
    InfoNCE   & MI     & 2  & 3    & 1.41E-03 & 1.52E-02 & 1.96E-03 & 8.42E-03 & 3.71E-02 \\
    \hline
    \end{tabular}
  \end{table}
  \begin{table}[]
    \centering
    \caption{Variance of KNIFE estimator and Gauss estimator in synthesized experiment~\label{table:knife-gauss-syn-var}}
    \begin{tabular}{llrrrrrrr}
    \hline
    Estimator & target & ND & task & 10       & 250      & 500      & 750      & 1000     \\
    \hline
    KNIFE     & CE     & 0  & 0    & 1.12E-03 & 1.18E-04 & 2.58E-04 & 1.22E-04 & 4.01E-04 \\
    KNIFE     & CE     & 1  & 0    & 1.32E-03 & 4.35E-04 & 1.67E-03 & 1.24E-03 & 1.60E-03 \\
    KNIFE     & CE     & 2  & 0    & 3.09E-04 & 3.47E-04 & 6.68E-04 & 5.01E-04 & 6.71E-04 \\
    KNIFE     & CE     & 0  & 1    & 1.20E-03 & 1.16E-03 & 9.20E-03 & 5.52E-04 & 3.86E-04 \\
    KNIFE     & CE     & 1  & 1    & 1.17E-03 & 1.78E-03 & 3.60E-04 & 7.84E-03 & 9.80E-04 \\
    KNIFE     & CE     & 2  & 1    & 2.01E-03 & 5.42E-03 & 7.77E-03 & 3.80E-03 & 2.97E-03 \\
    KNIFE     & CE     & 0  & 2    & 3.69E-02 & 5.77E-02 & 9.28E-02 & 3.86E-02 & 5.32E-02 \\
    KNIFE     & CE     & 1  & 2    & 2.40E-01 & 3.58E-02 & 8.50E-02 & 3.03E-02 & 2.33E-02 \\
    KNIFE     & CE     & 2  & 2    & 6.35E-02 & 1.23E-01 & 7.71E-02 & 7.32E-02 & 3.80E-02 \\
    KNIFE     & CE     & 0  & 3    & 9.86E-02 & 3.58E-02 & 1.39E-01 & 1.94E-02 & 1.64E-03 \\
    KNIFE     & CE     & 1  & 3    & 2.87E-01 & 9.47E-03 & 8.61E-02 & 2.80E-02 & 4.58E-03 \\
    KNIFE     & CE     & 2  & 3    & 1.16E-01 & 4.20E-02 & 1.48E-02 & 1.93E-02 & 6.58E-03 \\
    Gauss     & CE     & 0  & 0    & 5.23E-02 & 9.38E-02 & 2.44E-04 & 1.32E-04 & 1.48E-04 \\
    Gauss     & CE     & 1  & 0    & 1.28E+00 & 8.70E-02 & 3.66E-02 & 1.67E-04 & 1.11E-03 \\
    Gauss     & CE     & 2  & 0    & 1.30E+00 & 1.49E-01 & 3.90E-02 & 2.25E-02 & 1.50E-02 \\
    Gauss     & CE     & 0  & 1    & 3.84E-02 & 8.59E-02 & 8.49E-02 & 2.33E-02 & 2.06E-03 \\
    Gauss     & CE     & 1  & 1    & 8.54E-01 & 6.73E-04 & 4.60E-02 & 3.88E-02 & 1.58E-02 \\
    Gauss     & CE     & 2  & 1    & 6.48E-02 & 1.10E-01 & 5.69E-02 & 3.83E-04 & 3.52E-03 \\
    Gauss     & CE     & 0  & 2    & 5.47E+02 & 5.21E+00 & 1.04E+00 & 3.84E-02 & 1.12E-01 \\
    Gauss     & CE     & 1  & 2    & 6.38E-01 & 2.74E-03 & 8.14E-03 & 3.02E+01 & 2.78E+01 \\
    Gauss     & CE     & 2  & 2    & 4.10E-01 & 4.20E+15 & 2.74E-02 & 5.96E-01 & 1.46E+06 \\
    Gauss     & CE     & 0  & 3    & 9.21E-01 & 1.03E+00 & 2.40E-01 & 1.83E+01 & 1.06E-01 \\
    Gauss     & CE     & 1  & 3    & 4.39E+01 & 3.58E-01 & 3.75E+00 & 3.49E-01 & 4.08E+00 \\
    Gauss     & CE     & 2  & 3    & 7.32E-01 & 4.87E-01 & 8.85E+01 & 1.84E+00 & 1.27E+00 \\
    \hline
    \end{tabular}
    \end{table}
\subsection{Variance of Model Performance\label{append-pra-var}}
In Table~\ref{table:model-var}, we show variance of $R^2$ values of models.
In Table~\ref{table:indicators-var}, we show variance fo indicators.
\begin{table}[]
  \centering
  \caption{Variance of model in practical experiment~\label{table:model-var}}
  \begin{tabular}{lrrrrrr}
  \hline
  Task              & EV & lgbforest & lgbtree& m5& extratree & xgboost \\
  \hline
  app-energy & 0.906                  & 1.19E-04                      & 3.56E-04                    & 2.60E-04               & 2.78E-05                      & 2.40E-04                    \\
  app-energy & 0.701                  & 1.29E-04                      & 1.24E-03                    & 6.67E-05               & 9.69E-03                      & 1.89E-02                    \\
  app-energy & 0.518                  & 1.23E-05                      & 8.28E-05                    & 5.50E-06               & 2.37E-03                      & 2.36E-05                    \\
  BJ-pm25      & 0.936                  & 2.33E-05                      & 9.70E-05                    & 3.16E-04               & 6.75E-05                      & 5.62E-04                    \\
  BJ-pm25      & 0.811                  & 4.10E-06                      & 1.42E-06                    & 7.47E-04               & 2.06E-03                      & 2.66E-06                    \\
  bike-hour         & 0.919                  & 1.58E-04                      & 1.76E-03                    & 4.01E-03               & 1.28E-03                      & 7.82E-05                    \\
  bike-hour         & 0.728                  & 1.21E-04                      & 1.38E-03                    & 2.24E-03               & 2.78E-03                      & 1.17E-04                    \\
  CT-slices         & 0.901                  & 1.14E-05                      & 6.88E-07                    & 1.12E-07               & 5.45E-08                      & 4.92E-07                    \\
  CT-slices         & 0.702                  & 1.45E-05                      & 3.81E-07                    & 5.60E-06               & 1.14E-07                      & 1.16E-06                    \\
  CT-slices         & 0.516                  & 1.67E-05                      & 1.14E-06                    & 3.65E-06               & 3.06E-07                      & 1.99E-06                    \\
  CT-slices         & 0.320                  & 1.54E-04                      & 4.85E-05                    & 5.88E-05               & 1.89E-04                      & 1.59E-04                    \\
  comp-stren    & 0.969                  & 4.10E-03                      & 1.17E-03                    & 3.91E-02               & 7.93E-04                      & 2.35E-03                    \\
  comp-stren    & 0.884                  & 5.35E-03                      & 4.38E-03                    & 9.22E-02               & 1.25E-03                      & 3.96E-03                    \\
  comp-stren    & 0.774                  & 5.91E-04                      & 6.02E-04                    & 7.49E-05               & 1.99E-03                      & 6.05E-03                    \\
  comp-stren    & 0.587                  & 1.61E-03                      & 2.12E-03                    & 7.29E-05               & 1.75E-03                      & 7.33E-03                    \\
  comp-stren    & 0.333                  & 4.57E-04                      & 2.82E-03                    & 1.55E-03               & 1.79E-02                      & 6.04E-03\\
  \hline
  \end{tabular}
  \end{table}

    \begin{table}[]
      \centering
      \caption{Variance of indicators in practical experiment~\label{table:indicators-var}}
      \begin{tabular}{rrlll}
      \hline
      Task              & EV & $R^2_{knifecp}$ & $R^2_{lmcp}$ & $R^2_{knifedp}$\\
      \hline
      app-energy & 0.906                  & 1.35E-05                     & 1.17E-04                     & 2.30E-05                  \\
      app-energy & 0.701                  & 1.22E-04                     & 1.18E-04                     & 4.14E-05                  \\
      app-energy & 0.518                  & 4.23E-05                     & 8.22E-05                     & 5.64E-05                  \\
      BJ-pm25      & 0.936                  & 4.38E-03                     & 6.83E-03                     & 3.32E-03                  \\
      BJ-pm25      & 0.811                  & 4.49E-03                     & 4.68E-03                     & 2.68E-03                  \\
      bike-hour         & 0.919                  & 9.56E-04                     & 2.69E-04                     & 1.03E-03                  \\
      bike-hour         & 0.728                  & 5.33E-05                     & 2.36E-04                     & 3.93E-05                  \\
      CT-slices         & 0.901                  & 4.21E-05                     & 2.95E-05                     & 1.87E-07                  \\
      CT-slices         & 0.702                  & 1.79E-03                     & 5.74E-06                     & 7.64E-06                  \\
      CT-slices         & 0.516                  & 2.11E-06                     & 2.70E-06                     & 3.09E-09                  \\
      CT-slices         & 0.320                  & 6.18E-07                     & 6.84E-06                     & 3.29E-07                  \\
      comp-stren    & 0.969                  & 5.87E-04                     & 3.06E-04                     & 3.59E-04                  \\
      comp-stren    & 0.884                  & 6.91E-04                     & 1.47E-03                     & 8.94E-04                  \\
      comp-stren    & 0.774                  & 6.26E-04                     & 4.33E-03                     & 4.27E-04                  \\
      comp-stren    & 0.587                  & 9.45E-04                     & 9.31E-04                     & 4.38E-04                  \\
      comp-stren    & 0.333                  & 7.40E-04                     & 6.15E-04                     & 3.35E-04  \\
      \hline        
      \end{tabular}
      \end{table}
\section{Libraries Used}
For our experiments, we built upon code from the following sources and packages.
\begin{itemize}
    \item InfoNCE\citep{AO18}, MINE\citep{MIB18}, CLUB\citep{PC20}, Gauss(DoE)\citep{DM20}, KNIFE\citep{GP22}, at github.com/g-pichler/knife \citep{GP22}
    \item LightGBM \citep{GK17} at github.com/microsoft/LightGBM
    \item Xgboost \citep{TC16} at github.com/dmlc/xgboost
    \item M5 Tree at github.com/smarie/python-m5p
    \item ExtraTree from scikit-learn at \citep{SL11} at github.com/scikit-learn/scikit-learn
\end{itemize}

\end{appendices}
\end{document}